\documentclass[journal]{IEEEtran}
\usepackage{cite}

\ifCLASSINFOpdf
  \usepackage[pdftex]{graphicx}
\else
\fi
\usepackage{amsmath}
\usepackage{amsfonts, amssymb, amsthm}
\usepackage{bbding}
\usepackage{mathrsfs}
\usepackage{stfloats}

\usepackage{booktabs}
\usepackage{multirow}
\usepackage{color}

\usepackage{tikz,xcolor,hyperref}

\definecolor{lime}{HTML}{A6CE39}
\DeclareRobustCommand{\orcidicon}{
    \begin{tikzpicture}
    \draw[lime, fill=lime] (0,0)
    circle [radius=0.16]
    node[white] {{\fontfamily{qag}\selectfont \tiny \.{I}D}};
    \draw[white, fill=white] (-0.0625,0.095)
    circle [radius=0.007];
    \end{tikzpicture}
    \hspace{-2mm}
}

\foreach \x in {A, ..., Z}{\expandafter\xdef\csname orcid\x\endcsname{\noexpand\href{https://orcid.org/\csname orcidauthor\x\endcsname}
			{\noexpand\orcidicon}}
}

\begin{document}
%
\title{Cross Modification Attention Based\\ Deliberation Model for Image Captioning}
%
%
%

\author{Zheng Lian\hspace{-1.5mm}\orcidA{},
        Yanan Zhang\hspace{-1.5mm}\orcidB{},
        Haichang Li\hspace{-1.5mm}\orcidC{},
        Rui Wang\hspace{-1.5mm}\orcidD{},~\IEEEmembership{Member,~IEEE,}
        Xiaohui Hu\hspace{-1.5mm}\orcidE{}
\thanks{Z. Lian and Y. Zhang are with the Science \& Technology on Integrated Information System Laboratory,
Institute of Software Chinese Academy of Sciences, Beijing 100190, China, and also with the
University of Chinese Academy of Sciences, Beijing 100049, China (e-mail: lianzheng2017@iscas.ac.cn; yanan2018@iscas.ac.cn).}
\thanks{H. Li, R. Wang and X. Hu are with the Science \& Technology on Integrated Information System Laboratory,
Institute of Software Chinese Academy of Sciences, Beijing 100190, China (e-mail: haichang@iscas.ac.cn;
wangrui@iscas.ac.cn; hxh@iscas.ac.cn).}}

\maketitle

\begin{abstract}

The conventional encoder-decoder framework for image captioning generally adopts a single-pass decoding process,
which predicts the target descriptive sentence word by word in temporal order. Despite the great success of this
framework, it still suffers from two serious disadvantages. Firstly, it is unable to correct the mistakes in the
predicted words, which may mislead the subsequent prediction and result in error accumulation problem. Secondly,
such a framework can only leverage the already generated words but not the possible future words, and thus lacks
the ability of global planning on linguistic information. To overcome these limitations, we explore a universal
two-pass decoding framework, where a single-pass decoding based model serving as the \emph{Drafting Model} first
generates a draft caption according to an input image, and a \emph{Deliberation Model} then performs the polishing
process to refine the draft caption to a better image description. Furthermore, inspired from the complementarity
between different modalities, we propose a novel Cross Modification Attention (CMA) module to enhance the semantic
expression of the image features and filter out error information from the draft captions. We integrate CMA with
the decoder of our \emph{Deliberation Model} and name it as Cross Modification Attention based Deliberation Model
(CMA-DM). We train our proposed framework by jointly optimizing all trainable components from scratch with a trade-off
coefficient. Experiments on MS COCO dataset demonstrate that our approach obtains significant improvements over
single-pass decoding baselines and achieves competitive performances compared with other state-of-the-art two-pass
decoding based methods.

\end{abstract}


%
\IEEEpeerreviewmaketitle

\section{Introduction}
%
%
%
%

\IEEEPARstart{I}{mage} captioning is one of the quintessential sequence generation tasks,
which aims to automatically generate natural-language descriptions for images
\cite{kulkarni2013babytalk,vinyals2016show,xu2015show,cho2015describing,li2017gla,anderson2018bottom,guo2019show,wu2020recall,xu2020multi,ben2021unpaired}.
This task is exceedingly challenging due to its strict requirements for exact recognition
of salient objects, clear comprehension of their relationship and fluent expression
describing the visual contents with human language. Image captioning is beneficial for a
multitude of practical applications. For example, it can make it possible for visually
impaired people to understand the rich visual world.

\begin{figure}[!tp]
    \begin{center}
        \includegraphics[width=3.5in]{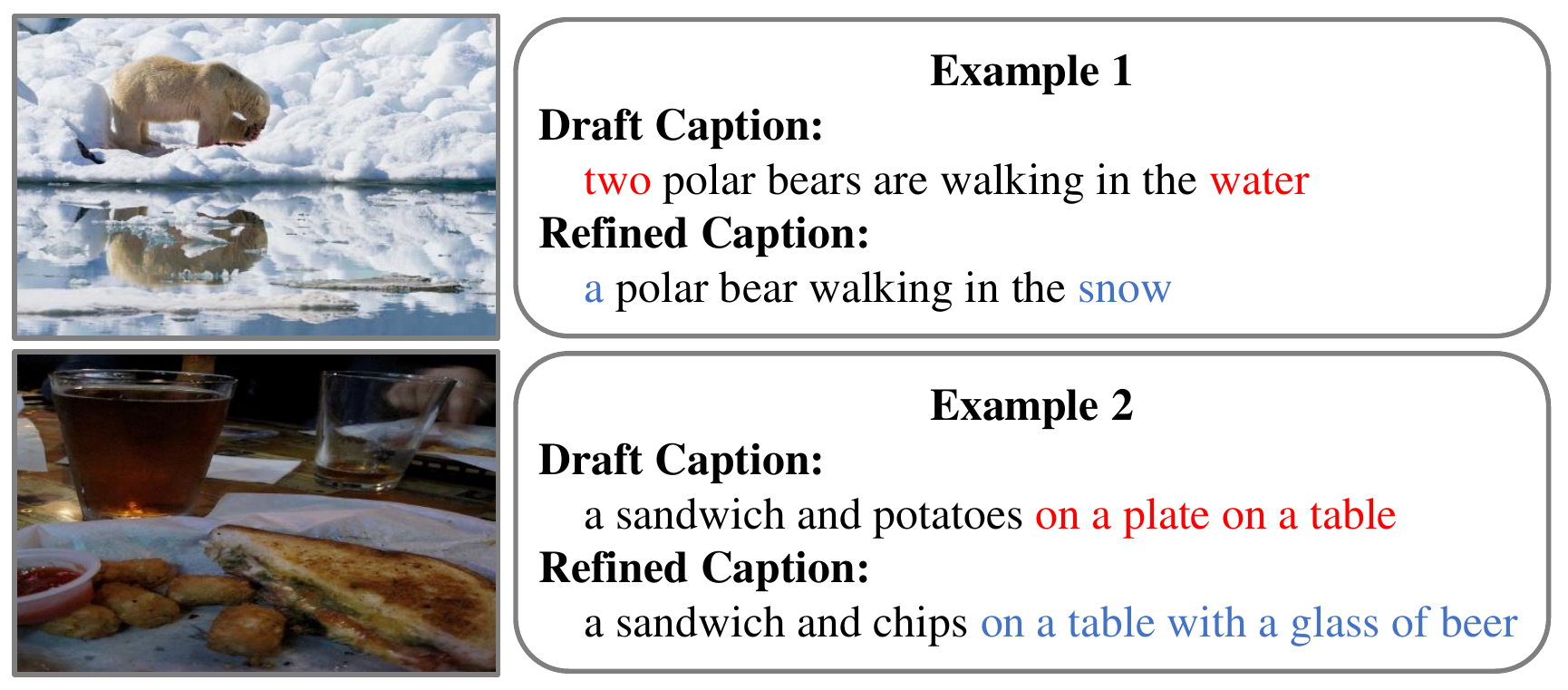}
    \end{center}
    \setlength{\abovecaptionskip}{0cm}
    \caption{Two examples produced by our two-pass decoding based method. Given an image,
    the \emph{Drafting Model} first generates a draft caption, and our CMA-DM then refines
    it to a better description in the polishing process. For the first example, our CMA-DM
    corrects ``two'' to ``a'' and avoids error accumulation. For the second example, our
    CMA-DM makes global planning based on the draft caption and generates a more readable
    descriptive sentence.}
    \label{fig_intro_examples}
\end{figure}

At present, the majority of image captioning models follow the conventional encoder-decoder framework \cite{vinyals2016show,
xu2015show,cho2015describing,li2017gla,anderson2018bottom,mao2014deep,yao2017boosting,yao2018exploring,huang2019attention},
where a Convolutional Neural Network (CNN) first encodes an input image into a sequence of regional features, and a Recurrent
Neural Network (RNN) then decodes the visual features to a descriptive sentence. Despite the great success of this framework,
it still suffers from two serious disadvantages. The first one is that such models adopt a regular single-pass decoding process,
which results in their inability to correct the mistakes in predicted words. Furthermore, mistakes made in early steps may lead
to error accumulation under the constraint of the language model. As shown in the first example of Fig. \ref{fig_intro_examples},
the single-pass decoding based model generates a wrong word ``two'' at the beginning of caption generation, which leads the model
to keep mining the visual cues related to the ``two polar bears''. Thus, another wrong word ``water'' is then generated to
describe the surrounding, which is a typical example of error accumulation. The second weakness is that when the decoder performs
to predict the \emph{t}-th word \emph{w$_{t}$}, it can only leverage the already generated words \emph{w$_{<t}$}, but not the
possible words \emph{w$_{>t}$} to be generated in the future. That is, the decoder in such a single-pass decoding framework lacks
the ability of global planning on linguistic information, which may degrade the semantic consistency and fluency of generated captions.
For the second example in Fig. \ref{fig_intro_examples}, the single-pass decoding based model generates two consecutive prepositional
phrases after the word ``potatoes''. Although either of them can express reasonable semantics in their local contexts, it makes the
sentence verbose and degrades its fluency to put them together for describing the image.

Deliberation is a rewarding behavior in human's daily life. It plays a predominant role in improving
the quality of reading and writing. When someone intends to produce high-level writings, they will
always make rough drafts first, and then refine their works by modifying the details. To overcome the
limitations of single-pass decoding based methods, in this paper, we explore a universal two-pass
decoding framework for image captioning, which is composed of two interrelated models: a
single-pass decoding based model serving as the \emph{Drafting Model} first produces a draft caption
according to an input image in the first-pass decoding process, and a \emph{Deliberation Model} then
performs the polishing process to refine the draft caption to a better descriptive sentence. In order
to enhance the correlation between the two models, a refining encoder is designed to participate in
both decoding processes. All trainable components are jointly optimized from scratch with a trade-off
coefficient to balance the performances of these two models.

Technically, the draft captions are completely derived from the input images and expected to approach
the ground truths as much as possible. That is, image features contain more original and reliable
information, while the draft captions carry more high level semantic information directly related to
human language. Inspired from the complementarity between visual and linguistic modalities, we propose
a novel Cross Modification Attention (CMA) module to perform mutual correction between the features of
input images and the draft captions. CMA first generates visual and textual context vectors by applying
the multi-head attention mechanism \cite{vaswani2017attention} to both visual and linguistic features,
and then distills the information beneficial to word generation from both context vectors with gated
operation. In order to prevent the visual features from being misinformed by wrong drafts while cross
modification, we further add a residual connection on the visual side of CMA. In this case, CMA
can not only correct the mistakes in the drafts, but also ensure the accuracy of visual features while
enhancing their semantic expression. Although CMA is designed as a structural extension of the multi-head
attention mechanism in this paper, it is also available for other attention modules, like additive attention \cite{xu2015show}.
We integrate CMA with the decoder of our \emph{Deliberation Model} and name it as Cross Modification
Attention based Deliberation Model (CMA-DM).

Within our two-pass decoding framework, CMA-DM shows powerful capability of error correction and global
planning. As described in the first example of Fig. \ref{fig_intro_examples}, our CMA-DM realizes the
only polar bear in the image and corrects the wrong word ``two'' to ``a'', which helps to predict the
accurate surrounding, ``snow''. For the second example, our CMA-DM performs global planning based on
the draft caption and refines it to a more semantic and readable description.

Main contributions of this paper are summarized as follows:
\begin{itemize}
\item We explore a universal two-pass decoding framework for image captioning, where a single-pass
decoding based captioning model serves as the \emph{Drafting Model} to produce a draft caption in the
first-pass decoding process, and a \emph{Deliberation Model} then performs the polishing process to refine
the draft caption to a better image description.
\item We integrate a well-designed Cross Modification Attention (CMA) module with the decoder of our
\emph{Deliberation Model}, denoted as CMA-DM, to perform mutual correction between the features of input
images and corresponding draft captions.
\item Experiments on MS COCO dataset \cite{lin2014microsoft} demonstrate that our approach obtains significant
improvements over single-pass decoding baselines and achieves competitive performances compared with other
state-of-the-art two-pass decoding based methods.
\end{itemize}




\section{Related Work}
\label{sec:related work}

\subsection{Single-Pass Decoding Based Image Captioning Models}
To date, most of the image captioning models have employed the conventional encoder-decoder
framework to generate the image descriptions through one-pass decoding process
\cite{vinyals2016show,xu2015show,cho2015describing,li2017gla,anderson2018bottom,mao2014deep,
yao2017boosting,yao2018exploring,huang2019attention}.
Vinyals \emph{et al.} \cite{vinyals2016show}
introduced an end-to-end captioning model (NIC), which adopts a CNN as the encoder and an LSTM \cite{hochreiter1997long}
instead of a vanilla RNN as the decoder.
Yao \emph{et al.} \cite{yao2017boosting} detected visual attributes from images by using Multiple
Instance Learning (MIL) for better image representations. Afterward, \cite{yao2018exploring}
built semantic/spatial directed graph on the detected regions with Graph Convolutional
Networks (GCNs) and integrated both semantic and spatial object relationships into the image
encoder for visual feature refinement. \cite{yang2019auto} represented the complex structural
layout of both image and sentence with the scene graph for generating more human-like captions.
Gu \emph{et al.} \cite{gu2017empirical} designed a language CNN model to capture the long-range dependencies
in sequences through multilayer convolutional networks. \cite{aneja2018convolutional} developed
a convolutional architecture for image captioning, where multiple layers of masked convolution
substitute for RNNs to decode the image features. \cite{qin2019look} proposed a Predict Forward
(PF) approach to predict the next two words in one time step against the exposure bias.


\subsection{Attention Mechanisms}
Attention mechanisms constitute a significant extension of the conventional encoder-decoder framework,
which allow the decoder to selectively focus on the most relevant features at each time step of sequence
generation.
Inspired by human intuition and the effective research in neural machine translation \cite{bahdanau2014neural},
Xu \emph{et al.} \cite{xu2015show} made the first attempt to incorporate the visual attention into image
captioning to dynamically attend to salient regions of the image while caption generation.
Anderson \emph{et al.} \cite{anderson2018bottom} introduced a combined bottom-up and top-down
attention mechanism, where the bottom-up attention leverages Faster R-CNN \cite{ren2015faster} to provide
image representations at the level of objects and the top-down attention is responsible for predicting the
weight distribution. Boosted Attention \cite{chen2018boosted} combined the stimulus-based attention with
top-down captioning attention to provide prior knowledge on salient regions. More recently, Huang \emph{et al.}
\cite{huang2019attention} designed an ``Attention on Attention'' (AoA) module to determine the relevance
between attention results and queries. \cite{qin2019look} proposed to take into account the previous attention
result when predicting the attention weights at current time. Significantly, these attention-based models
also adopt the single-pass decoding based framework and are able to work as the \emph{Drafting Model} as well.


\subsection{Deliberation Networks}
Deliberation is aimed at polishing the existing results for further improvement
\cite{xia2017deliberation,hu2020deliberation,hu2021transformer}. Review Net \cite{yang2016review} was a rudiment
of the deliberation network for image captioning, which outputs a thought vector after each review step to
capture the global properties in a compact vector representation. Motivated by the great progress on deliberation
networks in neural machine translation \cite{xia2017deliberation}, Gao \emph{et al.} \cite{gao2019deliberate}
presented a Deliberate Residual Attention Network (DA), which consists of two residual-based attention layers.
The first-pass residual-based attention layer generates raw context vectors, and then the second-pass deliberate
residual-based attention layer refines them for generating better captions. Although both methods involve the
deliberation process, they still stagnate in the single-pass decoding captioning framework, where they can only
leverage the generated words but not the future words at each time step of word generation.

Later on, Sammani \emph{et al.} \cite{sammani2019look} introduced a Modification Network (MN) to modify existing
captions from a given framework by modeling the residual information. MN uses a Deep Averaging Network (DAN) to
encode features of the existing captions into a fixed size representation and two separate LSTMs for attention
and language modeling. Latterly, \cite{sammani2020show} proposed a caption-editing model to perform iterative
adaptive refinement of an existing caption. 
The main advantage of these two methods is that they take the complete raw caption into consideration when predicting
each word. However, these two models are not jointly trained with the base captioning model.

Our work is related to ruminant decoder (RD) \cite{guo2019show}. It performs multi-pass
decoding process to produce a more comprehensive caption, which is derived by refining the raw caption generated
from the base model. There are three significant differences between RD and our proposed framework. The first one
is that the ability for RD to correct mistakes in the draft captions mainly comes from the supervision information
in the training process, while we design a novel Cross Modification Attention (CMA) module for explicit mutual
correction between visual and linguistic features. The second difference lies in the structure of captioning frameworks.
RD takes the hidden states of base decoder as input to the ruminant decoder, while our framework leverages the features
of the refining encoder to enhance the contact between \emph{Drafting Model} and \emph{Deliberation Model}. The
third difference lies in the training algorithms. RD adopts separately training method for cross entropy loss, and
performs reinforcement learning by alternately optimizing the base and ruminant decoders. We train our framework
by jointly optimizing all trainable components from scratch with a trade-off coefficient, which is a fairly simple
but effective training strategy to achieve better performance.

\section{Methodology}
\label{sec:method}
In this section, we focus on discussing our two-pass decoding framework for image captioning. To achieve
a clear explanation of the details, we first introduce the Cross Modification Attention (CMA) module and then
make an elaborate introduction to the overall framework along with each of its components, especially the Cross
Modification Attention-based Deliberation Model (CMA-DM). Finally, we describe our joint optimization algorithm
for both training stages with cross entropy loss and reinforcement learning.

\subsection{Cross Modification Attention}
\label{sec:cma}
\begin{figure}[!t]
    \begin{center}
        \includegraphics[width=3.5in]{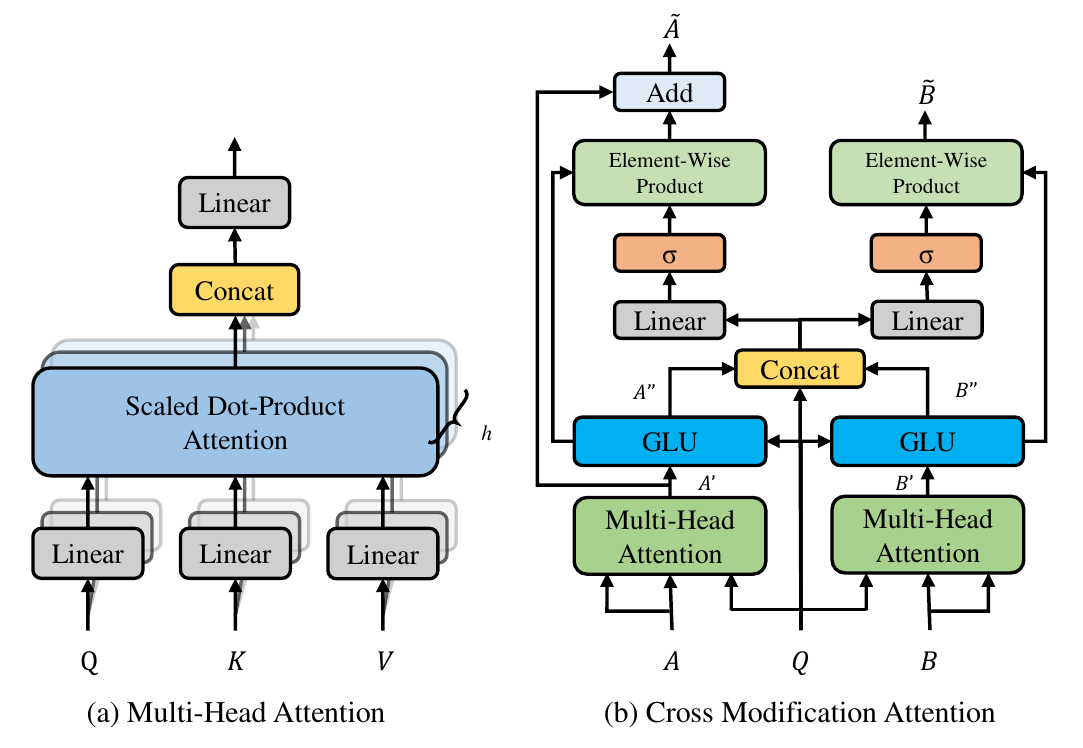}
    \end{center}
    \setlength{\abovecaptionskip}{-0.5cm}
    \caption{Multi-Head Attention and Cross Modification Attention. (a) Multi-head attention operates
    on a single set of key-value pairs and outputs the context vector by applying \emph{h} parallel
    scaled dot-product attention modules. (b) CMA operates on both visual and linguistic features and
    generates modified context vectors using gated operation and residual connection.}
    \label{fig_method_attentions}
\end{figure}

Multi-Head Attention $f_{mh-att}(Q, K, V)$ \cite{vaswani2017attention} integrates several parallel Scaled Dot-Product
Attention layers to jointly attend to information from different representation subspaces. It first
projects the queries, keys and values (denoted by $Q, K, V$ respectively) $h$ times with
different learnable linear transformation layers, and then applies the Scaled Dot-Product Attention
$f_{dot-att}(Q, K, V)$ to each group of the projected features concurrently. After that, the
attended results are concatenated and projected by one more linear transformation layer to form the
final context vector. Here, we assume that the channel dimensions of $Q, K, V$ are respectively
$d_q, d_k, d_v$. The Multi-Head Attention, as depicted in Fig. \ref{fig_method_attentions} (a), can be
formulated as follows:
\begin{align}
  f_{mh-att}(Q, K, V)&=Concat(head_1, \ldots, head_h)W^O \notag \\
  head_i&=f_{dot-att}(Q_i, K_i, V_i) \notag \\
  f_{dot-att}(Q_i, K_i, V_i)&=softmax(\frac{Q_iK_i^\mathsf{T}}{\sqrt{d_k}})V_i \notag \\
  Q_i=QW_i^Q, K_i&=KW_i^K, V_i=VW_i^V,
\end{align}
where $W_{i}^{Q} \in \mathbb{R}^{d_{q} \times d_{c}}$, $W_{i}^{K} \in \mathbb{R}^{d_{k} \times d_{c}}$,
$W_{i}^{V} \in \mathbb{R}^{d_{v} \times d_{c'}}$, $W^{O} \in \mathbb{R}^{hd_{c'} \times d_{o}}$ are
linear transformation matrices; $d_{c}$ is the common dimension of the projected queries and keys;
$d_{c'}$ and $d_{o}$ are the dimensions of the attended results and the final context vector.

\begin{figure*}[!tp]
    \begin{center}
        \includegraphics[width=5.5in]{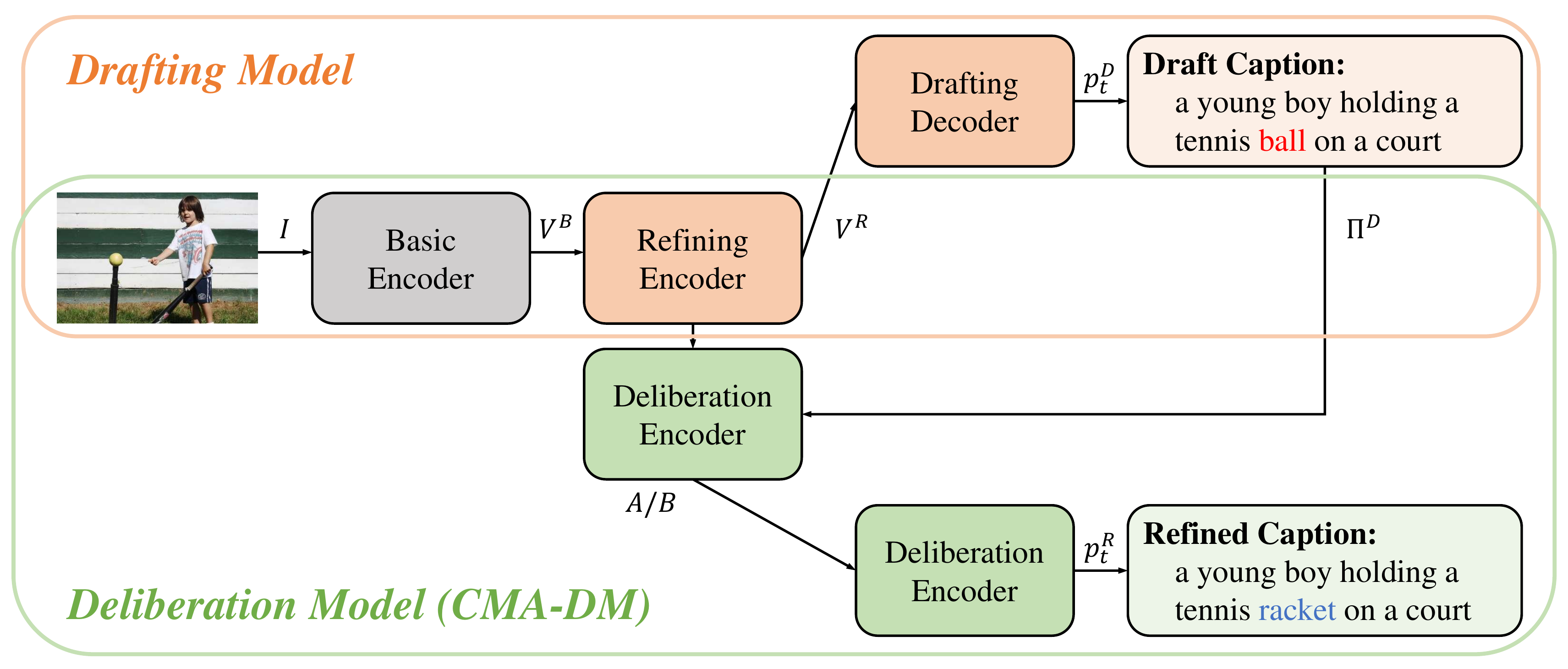}
    \end{center}
    \setlength{\abovecaptionskip}{0cm}
    \caption{Overview of our two-pass decoding framework for image captioning. It consists of two interrelated
    models: the \emph{Drafting Model} adopts the conventional encoder-decoder framework and generates a draft
    caption according to an given image in the first-pass decoding process; the \emph{Deliberation Model} then
    performs the polishing process to refine the draft caption to a better image description. Note that as the
    refining encoder is not a necessary structure for every single-pass decoding based model, we will add one
    to it when necessary to adapt to the architecture of our framework.}
    \label{fig_method_overall}
\end{figure*}

Such an attention module can only capture the query-related information from a single set of
key-value pairs. As it is necessary for our framework to leverage both visual and linguistic features
in the polishing process, we propose a Cross Modification Attention (CMA) module (as shown
in Fig. \ref{fig_method_attentions} (b)) to perform mutual correction between the features of images and draft captions. CMA
operates on some queries $Q \in \mathbb{R}^{k \times d_{q}}$, visual features $A \in \mathbb{R}^{n \times d_{a}}$
and linguistic features $B \in \mathbb{R}^{l \times d_{b}}$, where $k, n, l$ are the numbers of vectors
in $Q, A, B$. It first produces visual and linguistic context vectors $A'$, $B'$ by adopting two individual
Multi-Head Attention modules:
\begin{align}
  A'&=f_{mh-att}^{A}(Q, A, A) \notag \\
  B'&=f_{mh-att}^{B}(Q, B, B),
\end{align}
where $A', B' \in \mathbb{R}^{k \times d_{o}}$. Then CMA filters out the information that is unrelated to
the queries from both context features by using Gated Linear Units (GLU) \cite{dauphin2017language}:
\begin{align}
  A''&=f_{glu}(Q, A') \notag \\
     &=([Q; A']W_{p}^{A} + b_{p}^{A}) \odot \sigma([Q; A']W_{g}^{A} + b_{g}^{A}) \notag \\
  B''&=f_{glu}(Q, B') \notag \\
     &=([Q; B']W_{p}^{B} + b_{p}^{B}) \odot \sigma([Q; B']W_{g}^{B} + b_{g}^{B}),
\end{align}
where [;] indicates concatenation, $W_{p}^{A}, W_{g}^{A}, W_{p}^{B}, W_{g}^{B} \in \mathbb{R}^{(d_{q} + d_{o}) \times d_{o}}$,
$b_{p}^{A}, b_{g}^{A}, b_{p}^{B}, b_{g}^{B} \in \mathbb{R}^{d_{o}}$, $\sigma$ denotes the sigmoid activation
function and $\odot$ is is element-wise multiplication.

After that, the CMA makes mutual correction between the both gated features. This operation is also conditioned
on the queries $Q$. As the draft captions sometimes carry error messages, they will make the visual context
lose some valuable information during modifying. To alleviate this problem, we further add a residual
connection between the modified and the original visual context vectors:
\begin{align}
  \widetilde{A}&=A'' \odot \sigma([Q; A''; B'']W_{c}^{A} + b_{c}^{A}) + A' \notag \\
  \widetilde{B}&=B'' \odot \sigma([Q; A''; B'']W_{c}^{B} + b_{c}^{B}),
\end{align}
where $W_{c}^{A}, W_{c}^{B} \in \mathbb{R}^{(d_{q} + 2d_{o}) \times d_{o}}$ and $b_{c}^{A}, b_{c}^{B} \in \mathbb{R}^{d_{o}}$.
Above all, we obtain the modified visual and linguistic features by using the CMA $f_{cma}(Q, A, B)$:
\begin{align}
  \widetilde{A}, \widetilde{B}=f_{cma}(Q, A, B).
\end{align}

\subsection{Overall Framework}

A complete overview of our two-pass decoding framework is shown in Fig. \ref{fig_method_overall}. It consists
of two interrelated models: the \emph{Drafting Model} and the Cross Modification Attention-based Deliberation
Model (CMA-DM). The \emph{Drafting Model} adopts the standard encoder-decoder framework and performs the
first-pass decoding process. It first encodes an input image into a set of feature vectors, and then outputs
a sequence of draft caption by decoding the visual representations. In the second-pass decoding process, the
CMA module distills the information beneficial to current word generation from both the image and the draft
caption and helps the \emph{Deliberation Model} generate the refined caption with
higher quality. With the help of CMA, the \emph{Deliberation Model} can effectively filter out the error
information in the draft captions and enhance the semantic expression of the image features. In general,
our CMA-DM not only has access to the ability of global planning like other deliberation models, but also can
correct mistakes in the draft captions through a powerful attention module. In the following sections, we will
describe each of the two models in detail.

\subsection{Drafting Model}
Within our proposed framework, any well-designed single-pass decoding based image captioning model can act as
the \emph{Drafting Model} to transform the given images into draft captions in the first-pass decoding process.
In particular, the \emph{Drafting Model} in our framework generally consists of three components: the basic
encoder, the refining encoder and the drafting decoder.

\textbf{Basic Encoder:} Given an input image $I$, the basic encoder first extracts a set of feature vectors
$V^{B} = \{v_{1}^{B}, v_{2}^{B}, \ldots, v_{n}^{B}\}$, $v_{i}^{B} \in \mathbb{R}^{d_{B}}$. More specifically,
the \emph{Drafting Model} mostly employs a pre-trained CNN \cite{simonyan2014very,he2016deep} or R-CNN
\cite{ren2015faster} based network as the basic encoder, whose parameters are freezed during all the training
process.
\begin{align}
  V^{B}=f_{bs-enc}(I),
\end{align}
where $f_{bs-enc}$ is the encoding function depending on the network structure of the basic encoder.

\textbf{Refining Encoder:} Instead of directly feeding the fixed features $V^{B}$ to the subsequent components,
our \emph{Drafting Model} prefers to use an additional refining encoder to refine the visual representations.
As not all single-pass decoding based models adopt this module, we will add one to it when necessary to adapt to our
framework.
\begin{align}
  V^{R}=f_{rf-enc}(V^{B}),
\end{align}
where $V^{R} \in \mathbb{R}^{n \times d_{R}}$ is the refined features; $f_{rf-enc}$ is the refining function,
which exists in various forms in different drafting models, such as linear transformation layers and attention
functions. Significantly, the refined features not only participate in the generation of draft captions, but
also the deliberation process.

\textbf{Drafting Decoder:} With the refined visual features $V^{R}$, the drafting decoder generates a sequence
of draft caption $W^{D} = \{w_{1}^{D}, w_{2}^{D}, \ldots, w_{l}^{D}\}$
, where $l$ is the
length of the draft caption. At the $t$-th step, the drafting decoder outputs a probability distribution
$p_{t}^{D}$ to predict the next word $w_{t}^{D}$:
\begin{align}
  w_{t}^{D} \sim p_{t}^{D} = f_{dra-dec}(V^{R}, w_{t-1}^{D}),
\end{align}
where $f_{dra-dec}$ is the decoding function depending on the network structure of the drafting decoder.

\subsection{Cross Modification Attention-Based Deliberation Model}

In the first-pass decoding process, the \emph{Drafting Model} transforms the given image $I$ into a sequence of
draft caption $W^{D}$ word by word in temporal order. Therefore, it has no chance to correct mistakes in the
predicted words. That is, the draft captions may contain some mistakes. Moreover, due to the lack of global
planning ability, the \emph{Drafting Model} cannot guarantee the semantic consistency and fluency of the draft
captions. To address these problems, we propose a Cross Modification Attention-based Deliberation Model (CMA-DM)
to polish the draft captions in the second-pass decoding process.

\begin{figure}[!tp]
    \begin{center}
        \includegraphics[width=3.5in]{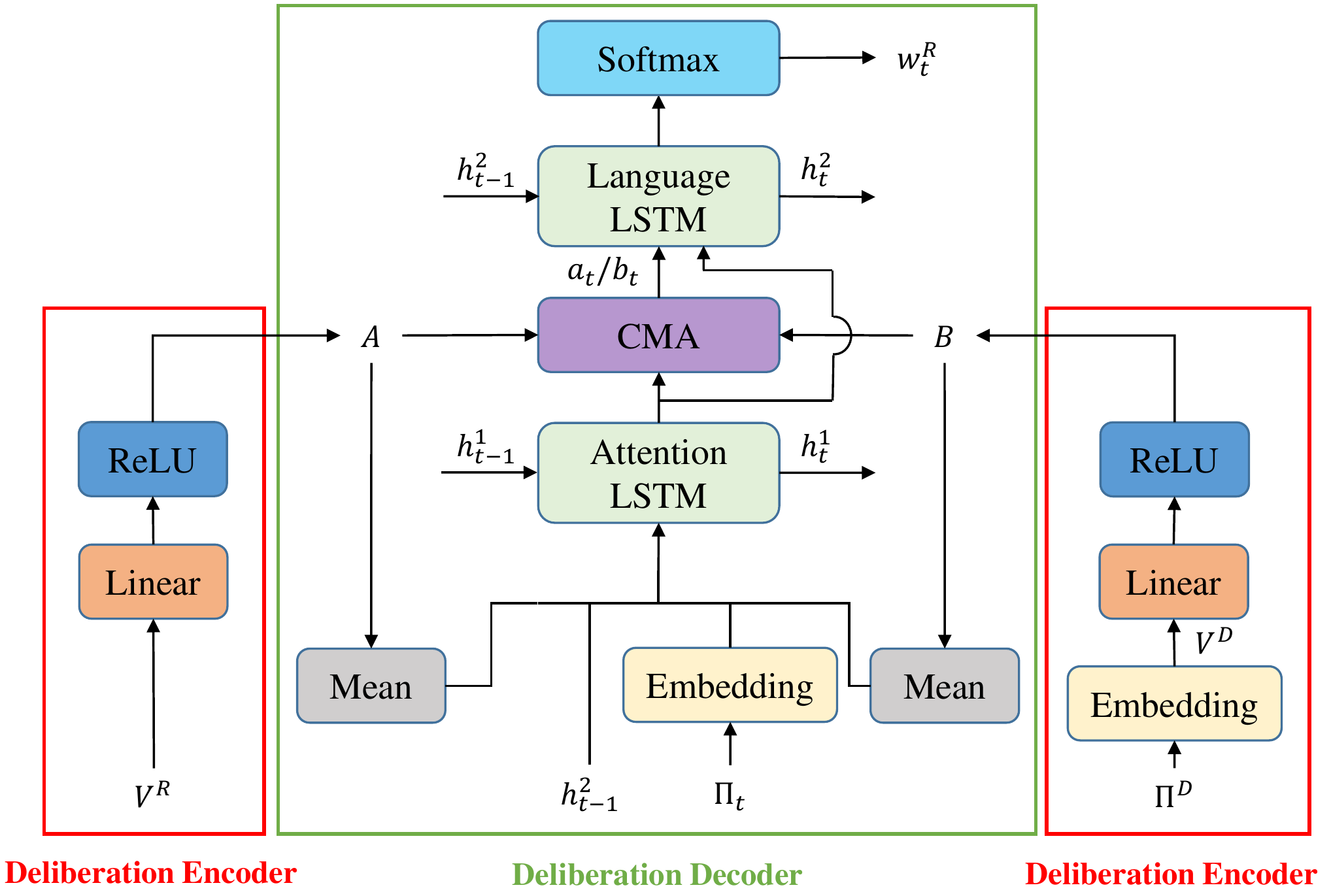}
    \end{center}
    \setlength{\abovecaptionskip}{0cm}
    \caption{Main structure of our CMA-DM. It consists of two components: the deliberation encoder projects
    both visual and linguistic representations into a new feature space; the deliberation decoder takes the
    projected features as input and refines the draft caption to a better image description. Note that the
    basic encoder and the refining encoder are not illustrated in this figure for clarity.}
    \label{fig_dm}
\end{figure}

As illustrated in Fig. \ref{fig_dm}, CMA-DM takes as input the refined visual features $V^{R}$ and the draft
caption $W^{D}$, and generates the refined caption $W^{R} = \{w_{1}^{R}, w_{2}^{R}, \ldots, w_{m}^{R}\}$, where
$m$ is the length of the refined caption. Our CMA-DM is composed of a deliberation encoder and a deliberation
decoder.

\textbf{Deliberation Encoder:} To realize the strategy of global planning, the deliberation encoder first
projects the one-hot encoding $\Pi^{D}$ of the input draft caption $W^{D}$ into dense vectors
$V^{D} = \{v_{1}^{D}, v_{2}^{D}, \ldots, v_{l}^{D}\}$, where $v_{i}^{D} \in \mathbb{R}^{d_{D}}$, by using
a single word embedding layer:
\begin{align}
  V^{D}=\Pi^{D}W_{e}^{D},
\end{align}
where $W_{e}^{D} \in \mathbb{R}^{|\Sigma| \times d_{D}}$ is a word embedding matrix for a vocabulary $\Sigma$,
and $V^{D} \in \mathbb{R}^{l \times d_{D}}$ is the dense feature vectors of the draft caption.

Rather than directly feeding the refined visual features $V^{R}$ and the dense word embeddings $V^{D}$ into the
deliberation decoder, we make a projection on both feature vectors with two individual linear transformation
layers followed by the Rectified Linear Unit (ReLU) \cite{glorot2011deep} activation function:
\begin{align}
  A&=ReLU(V^{R}W_{d}^{A}+b_{d}^{A}) \notag \\
  B&=ReLU(V^{D}W_{d}^{B}+b_{d}^{B}),
\end{align}
where $W_{d}^{A} \in \mathbb{R}^{d_{R} \times d_{D}}$, $W_{d}^{B} \in \mathbb{R}^{d_{D} \times d_{D}}$, and
$b_{d}^{A}, b_{d}^{B} \in \mathbb{R}^{d_{D}}$.

\textbf{Deliberation Decoder:} We adopt the widely-used top-down architecture from \cite{anderson2018bottom} as
the general structure of our deliberation decoder, as depicted in Fig. \ref{fig_dm}. It is composed of an
attention LSTM, a Cross Modification Attention module and a language LSTM.

In order to provide a strong query with the knowledge of global semantic information, we feed the mean-pooled
features of $A$ and $B$ to the attention LSTM. Other input vectors contains: the embedding vector of the input
word at current time step $\Pi_{t}$ and the last hidden state of the language LSTM $h_{t-1}^{2}$. That is, the
input of the attention LSTM at $t$-th step $x_{t}^{1}$ is given by:
\begin{align}
  x_{t}^{1}&=[\Pi_{t}W_{e}^{1}; \bar{A}; \bar{B}; h_{t-1}^{2}] \notag \\
  \bar{A}&=\textstyle\sum_{i=1}^{n}A_{i} \notag \\
  \bar{B}&=\textstyle\sum_{i=1}^{l}B_{i},
\end{align}
where $\Pi_{t}$ is the one-hot encoding vector of the input word and $W_{e}^{1} \in \mathbb{R}^{|\Sigma| \times d_{D}}$
is another word embedding matrix different from $W_{e}^{D}$.

Then, we consider the current hidden state of the attention LSTM as the query $Q$ to compute the context vectors
from both visual and linguistic features through CMA:
\begin{align}
  h_{t}^{1}, c_{t}^{1}&=LSTM^{1}(x_{t}^{1}, h_{t-1}^{1}, c_{t-1}^{1}) \notag \\
  a_{t}, b_{t}&=f_{cma}(h_{t}^{1}, A, B),
\end{align}
where $a_{t}, b_{t} \in \mathbb{R}^{d_{D}}$ are respectively the visual context vectors and the linguistic context
vectors.

Finally, we feed the following inputs to the language LSTM to get the probability distribution of each word in
the refined caption: both context vectors and the current hidden state of the attention LSTM:
\begin{align}
  x_{t}^{2}&=[h_{t}^{1}; a_{t}; b_{t}] \notag \\
  h_{t}^{2}, c_{t}^{2}&=LSTM^{2}(x_{t}^{2}, h_{t-1}^{2}, c_{t-1}^{2}) \notag \\
  w_{t}^{R} \sim p_{t}^{R}&=softmax(h_{t}^{2}).
\end{align}

\subsection{Joint Optimization Algorithm}
During all the training process, the basic encoder is kept fixed, and the other components are jointly optimized from
scratch with a trade-off coefficient. Denote the refining encoder as $\mathcal{E}_{r}(\theta_{r})$, the drafting decoder
as $\mathcal{D}_{d1}(\theta_{d1})$, the deliberation encoder as $\mathcal{E}_{d}(\theta_{d})$ and the deliberation
decoder $\mathcal{D}_{2}(\theta_{d2})$ parameterized with $\theta_{r}$, $\theta_{d1}$, $\theta_{d}$ and $\theta_{d2}$,
respectively. Obviously, the \emph{Drafting Model} $\mathcal{M}_{1}(\theta_{1})$ is accompanied by the parameters
$\theta_{1} = \theta_{r} \cup \theta_{d1}$, and the \emph{Deliberation Model} $\mathcal{M}_{2}(\theta_{2})$ appears with
the parameter $\theta_{2} = \theta_{r} \cup \theta_{d} \cup \theta_{d2}$. Like previous methods, we first pre-train our
framework with cross entropy (XE) loss and then directly optimize the non-differentiable metrics with Self-Critical
Sequence Training (SCST) \cite{rennie2017self}.

Given an input image $I$ and a target ground-truth sequence $W = w_{1:T} = (w_{1}, \ldots, w_{T})$, we first optimize
XE loss for both models in our framework with a trade-off coefficient $\lambda_{XE}$:
\begin{align}
    \mathcal{L}_{XE}(\theta_{1})&=-\textstyle\sum_{t=1}^{T}log(p_{\theta_{1}}(w_{t}|w_{1:(t-1)},I)) \notag \\
    \mathcal{L}_{XE}(\theta_{2})&=-\textstyle\sum_{t=1}^{T}log(p_{\theta_{2}}(w_{t}|w_{1:(t-1)},I,W^{D})) \notag \\
    \mathcal{L}_{XE}(\theta_{1}, \theta_{2})&=\mathcal{L}_{XE}(\theta_{1})+\lambda_{XE}\mathcal{L}_{XE}(\theta_{2}),
\end{align}
where $W^{D}$ is the draft caption generated by $\mathcal{M}_{1}$ in the first-pass decoding process.

Then we apply the standard SCST algorithm to each of the models. Similar to XE training stage, we employ another
trade-off coefficient $\lambda_{RL}$ to realize joint optimization for our framework:
\begin{align}
    \mathcal{L}_{RL}(\theta_{1})&=-\mathbb{E}_{w_{s}^{D} \sim p_{\theta_{1}}}[r(w_{s}^{D})] \notag \\
    \mathcal{L}_{RL}(\theta_{2})&=-\mathbb{E}_{w_{s}^{R} \sim p_{\theta_{2}}}[r(w_{s}^{R})] \notag \\
    \mathcal{L}_{RL}(\theta_{1}, \theta_{2})&=\mathcal{L}_{RL}(\theta_{1}) + \lambda_{RL} \mathcal{L}_{RL}(\theta_{2}),
\end{align}
where $w_{s}^{D}$ and $w_{s}^{R}$ are sentences sampled from the predicted word distribution produced by $\mathcal{M}_{1}$
and $\mathcal{M}_{2}$, respectively; the reward $r(\cdot)$ uses CIDEr score in this paper.
The gradients of each model can be approximated using a single Monte-Carlo sample from
$p_{\theta_{1}}$ or $p_{\theta_{2}}$:
\begin{align}
    \nabla_{\theta_{1}}\mathcal{L}_{RL}(\theta_{1}) &\approx -(r(w_{s}^{D}) - b_{1})\nabla_{\theta_{1}}logp_{\theta_{1}}(w_{s}^{D}) \notag \\
    \nabla_{\theta_{2}}\mathcal{L}_{RL}(\theta_{2}) &\approx -(r(w_{s}^{R}) - b_{2})\nabla_{\theta_{2}}logp_{\theta_{2}}(w_{s}^{R}),
\end{align}
where $b_{1}$ and $b_{2}$ are the baselines under the greedy decoding algorithm.

\section{Experimental Setup}
\label{sec:exp_setup}

\subsection{Datasets and Settings}
\textbf{Datasets:} All of our experiments are conducted on the most popular MS COCO dataset \cite{lin2014microsoft}.
MS COCO dataset contains a total of 123,287 images, including 82,783 training images, 40,504 validation images and
40,775 testing images. Each image is equipped with at least 5 human-annotated captions, and in this paper, we select
5 ground-truth sentences per image for performance evaluation like most studies. It is worth noting that the official
testing set does not provide the annotations corresponding to the images, so we can only make the evaluation over the
testing set through the online MS COCO testing server. Following previous work, we adopt the widely-used ``Karpathy''
split\footnote{[Online]. Available: https://cs.stanford.edu/people/karpathy/deepimagesent/} from \cite{karpathy2015deep}
for the offline evaluation, where 113,287 images are used for training, 5,000 images for validation and 5,000 images
for testing.

Visual Genome \cite{krishna2016visual} is a large-scale database containing over 108K images with dense annotations,
where each image has an average of 35 objects, 26 attributes and 21 pairwise relationships between objects. To
sufficiently describe all the contents and interactions in an image, Visual Genome provides on average 50 region
descriptions per image, and each description consists of 1 to 16 words. In this paper, we adopt the Faster R-CNN \cite{ren2015faster}
pre-trained on Visual Genome dataset as the basic encoder in the \emph{Drafting Model} to detect the bottom-up image features.

\textbf{Data Preprocessing:} We first convert all sentences to lower case and count the occurrences
of each word. In this paper, we only keep the words which appear more than 5 times, and replace the
discarded words with the ``UNK'' token. Then we truncate the captions to no longer than 16 words, and
those captions with less than 16 words are padded with the ``PAD'' token. Considering these two additional
tokens, we obtain the final vocabulary of 10,369 unique words. We further add another two ``PAD'' tokens
at both sides of each caption to symbolize the beginning and the end of it. By doing this, we finally
obtain the processed captions with a fixed length of 18 for training.

\textbf{Evaluation Metrics:} We mainly make the performance evaluation of our proposed approach on 5 different metrics:
BLEU (including BLEU-1,2,3,4) \cite{papineni2002bleu}, METEOR \cite{denkowski2014meteor}, ROUGE\_L \cite{lin2004rouge},
CIDEr \cite{vedantam2015cider} and SPICE \cite{anderson2016spice}, denoted as B\@1,2,3,4, M, R, C and S. These metrics
measure n-gram overlap degree between
generated sentences and reference sentences from different aspects. We also evaluate our CMA-DM with Word Mover's Distance
(WMD) \cite{kusner2015word} (denoted as W), which is first proposed to measure document distance by calculating Earth Mover's Distance
based on \emph{word2vec} embeddings of the words. All these metrics are computed with the publicly released MS COCO caption
evaluation tool\footnote{[Online]. Available: https://github.com/tylin/coco-caption/}.

\subsection{Drafting Models}
\label{exp_setup_drafting_models}
To convincingly demonstrate the powerful error-correcting capability of CMA-DM and the universality of our
proposed two-pass end-to-end framework for image captioning, we conduct several experiments by employing
three different single-pass decoding based models as the \emph{Drafting Model}. We make a careful comparison
between the following models and the corresponding CMA-DM in Section \ref{sec:exp_results}.

\textbf{a) Att2in:} Att2in model is a slightly modified version of the soft attention model \cite{xu2015show},
which feeds the attention-derived image features only to the cell node of the LSTM. Rather than using ResNet-101 in \cite{rennie2017self},
we adopt the Faster R-CNN pre-trained on Visual Genome dataset as the basic encoder for better visual representations.
To adapt to the structure of our proposed framework, we further employ a linear transformation layer followed
by an ReLU activation function as the refining encoder to refine the detected image features.

\textbf{b) Up-Down:} Up-Down \cite{anderson2018bottom} model is a widely-used architecture for image captioning, on which
many models are designed, including our CMA-DM. It first encodes the given images into
object-level features by using a Faster R-CNN pre-trained on Visual Genome dataset, and then
decodes the visual representations through two LSTMs with visual attention mechanism. Similar to
Att2in model, we also project the object-level features with a linear transformation layer
followed by ReLU before feeding them into the drafting decoder.

\textbf{c) AoANet:} AoANet \cite{huang2019attention} is the \emph{Base/Drafting Model} of the current state-of-the-art two-pass
decoding method \cite{sammani2020show}. It takes as input the detected features from Up-Down model, and then refines
them by using an AoA module based encoder. As AoANet conforms to the architecture of our \emph{Drafting Model},
we directly treat the pre-trained Faster R-CNN and the AoA encoder as the basic encoder
and the refining encoder respectively, without further adding extra layers. In this paper, we
adopt AoANet as our main \emph{Drafting Model} and most of the experimental results reported in Section \ref{sec:exp_results}
are based on it.

\begin{table*}[tp]
    \begin{center}
        \caption{Performance Comparisons of Our Method with Three Single-Pass Decoding Baselines on MS COCO ``Karpathy''
        Test Split. All Values are Reported as Percentage (\%). $^{\sharp}$ Indicates an Adaptive Version to the Structure
        of Our Drafting Model, and $^{\ast}$ Indicates the Results Obtained from the Publicly Available Pre-Trained
        Model. (-) Indicates an Unknown Metric.}
        \label{tab_baselines}
        \setlength{\tabcolsep}{1.10mm}{
        \begin{tabular}{cccccccccccccccccccc}
            \toprule
            \multicolumn{1}{c}{} &{} &\multicolumn{9}{c}{Cross Entropy Loss} &\multicolumn{9}{c}{Reinforcement Learning} \\
            \cmidrule(lr){3-11}
            \cmidrule(lr){12-20}
            \multicolumn{1}{c}{Group} &{Method} &{B@1} &{B@2} &{B@3} &{B@4} &{M} &{R} &{C} &{S} &{W} &{B@1} &{B@2} &{B@3} &{B@4} &{M} &{R} &{C} &{S} &{W} \\
            \midrule
            \midrule
            \multirow{4}{*}{1}
            &\multicolumn{1}{|l|}{Att2in \cite{rennie2017self}}
            & -    & -    & -    & 31.3 & 26.0 & 54.3 & 101.3 & -    &\multicolumn{1}{c|}{-}
            & -    & -    & -    & 33.3 & 26.3 & 55.3 & 111.4 & -    & -    \\
            &\multicolumn{1}{|l|}{Att2in$^{\sharp}$}
            & 76.6 & 60.6 & 46.9 & 35.8 & 27.1 & 56.5 & 112.2 & 20.4 &\multicolumn{1}{c|}{24.4}
            & 78.2 & 62.4 & 48.1 & 36.8 & 27.5 & 57.2 & 118.4 & 20.7 & 25.1 \\
            &\multicolumn{1}{|l|}{Att2in(Drafting)}
            & 76.3 & 60.0 & 46.2 & 35.5 & 27.1 & 56.2 & 111.7 & 20.1 &\multicolumn{1}{c|}{24.4}
            & 77.4 & 62.0 & 47.3 & 36.5 & 27.4 & 57.1 & 114.3 & 20.4 & 24.7 \\
            &\multicolumn{1}{|l|}{Att2in + CMA-DM}
            &\textbf{78.3} &\textbf{62.5} &\textbf{47.7} &\textbf{35.9} &\textbf{27.2} &\textbf{56.9} &\textbf{118.7} &\textbf{20.7} &\multicolumn{1}{c|}{\textbf{24.7}}
            &\textbf{79.5} &\textbf{63.7} &\textbf{48.8} &\textbf{37.1} &\textbf{27.7} &\textbf{57.6} &\textbf{121.9} &\textbf{21.1} &\textbf{25.2} \\ \midrule
            \multirow{4}{*}{2}
            &\multicolumn{1}{|l|}{Up-Down \cite{anderson2018bottom}}
            & 77.2 & -    & -    & 36.2 & 27.0 & 56.4 & 113.5 & 20.3 &\multicolumn{1}{c|}{-}
            & 79.8 & -    & -    & 36.3 & 27.7 & 56.9 & 120.1 & 21.4 & -    \\
            &\multicolumn{1}{|l|}{Up-Down$^{\sharp}$}
            & 76.5 & 60.5 & 46.8 & 36.2 &\textbf{27.8} & 56.6 & 113.7 & 20.6 &\multicolumn{1}{c|}{25.4}
            & 79.7 & 64.0 & 49.3 & 37.4 & 28.1 & 57.9 & 124.1 & 21.3 & 25.4 \\
            &\multicolumn{1}{|l|}{Up-Down(Drafting)}
            & 76.0 & 59.9 & 46.4 & 35.9 & 27.7 & 56.3 & 113.0 & 20.7 &\multicolumn{1}{c|}{25.3}
            & 79.4 & 63.7 & 49.0 & 37.2 & 27.9 & 57.9 & 122.5 & 21.2 & 25.4 \\
            &\multicolumn{1}{|l|}{Up-Down + CMA-DM}
            &\textbf{78.0} &\textbf{62.5} &\textbf{48.6} &\textbf{37.5} &\textbf{27.8} &\textbf{57.2} &\textbf{115.1} &\textbf{20.9} &\multicolumn{1}{c|}{\textbf{25.6}}
            &\textbf{80.1} &\textbf{64.5} &\textbf{50.0} &\textbf{38.0} &\textbf{28.3} &\textbf{58.2} &\textbf{125.0} &\textbf{21.7} &\textbf{26.2}    \\ \midrule
            \multirow{4}{*}{3}
            &\multicolumn{1}{|l|}{AoANet \cite{huang2019attention}}
            & 77.4 & -    & -    & 37.2 & 28.4 & 57.5 & 119.8 & 21.3 &\multicolumn{1}{c|}{-}
            & 80.2 & -    & -    & 38.9 &\textbf{29.2} & 58.8 &\textbf{129.8} & 22.4 & -    \\
            &\multicolumn{1}{|l|}{AoANet$^{\ast}$}
            & 77.3 & 61.6 & 47.9 & 36.9 &\textbf{28.5} & 57.3 & 118.4 &\textbf{21.7} &\multicolumn{1}{c|}{-}
            & 80.5 & 65.2 & 51.0 & 39.1 & 29.0 &\textbf{58.9} & 128.9 &\textbf{22.7} & -    \\
            &\multicolumn{1}{|l|}{AoANet(Drafting)}
            & 76.9 & 60.9 & 47.4 & 36.8 & 28.2 & 57.0 & 116.2 & 21.0 &\multicolumn{1}{c|}{26.0}
            & 80.4 & 65.0 & 50.8 & 39.0 & 28.9 & 58.8 & 128.3 & 22.4 & 26.8 \\
            &\multicolumn{1}{|l|}{AoANet + CMA-DM}
            &\textbf{78.6} &\textbf{63.6} &\textbf{49.9} &\textbf{38.7} &28.4 &\textbf{58.2} &\textbf{120.2} &21.4 &\multicolumn{1}{c|}{\textbf{26.2}}
            &\textbf{80.6} &\textbf{65.4} &\textbf{51.2} &\textbf{39.2} & 29.0 &\textbf{58.9} & 129.0 & 22.6 &\textbf{26.9} \\ \bottomrule
        \end{tabular}}
    \end{center}
\end{table*}

\subsection{Implement Details}
We integrate each of the aforementioned drafting models with our CMA-DM to build the overall
two-pass decoding framework.

\textbf{Basic Encoder:} As described in Section \ref{exp_setup_drafting_models}, we adopt the Faster
R-CNN pre-trained on Visual Genome dataset as the basic encoder in all experiments. We choose the
``adaptive" version of the bottom-up features extracted by the pre-trained Faster R-CNN, which
sets the class detection confidence threshold to 0.2 and detects 10\textasciitilde100 regions per
image. More specifically, the dimension of the extracted features is 2,048.

\textbf{Refining Encoder:} For Att2in model and Up-Down model, the refining encoders of them project
the extracted features to dimensions of 512 and 1,024, respectively. The refining module of AoANet
integrates the self-attentive multi-head attention mechanism with the AoA module, which is stacked
for 6 times and outputs the refined visual features with a dimension of 1,024.

\textbf{Drafting Decoder:} The original Att2in model and Up-Down model serve as the drafting decoder
to perform the first-pass decoding process. For Att2in model, we set the dimensions of word embeddings
and LSTM hidden states to 512. For Up-Down model, we set the dimensions of word embeddings and hidden
states in each LSTM to 1,000, and the number of hidden units in the attention layer to 512. As for the
decoder of AoANet, we set the dimensions of word embeddings and LSTM hidden states to 1,024. We employ
8 parallel attention heads in AoA module and set the dimension of the attended results for each head
to 128. For all drafting models, we set the maximum length of the generated draft captions to 16.

\textbf{CMA-DM:} Before feeding the one-hot encodings of draft captions into the word embedding layer,
we first pad the draft captions to 18 words with ``PAD'' tokens, which also contain the beginning and end
symbols. We set the dimensions of both word embedding layers (one for draft captions, the other one for
current input word) to 1,024. The deliberation encoder projects the refined visual features and the word
embeddings of draft captions into the same dimension of 1,024, which is also the dimensions of the global
visual/linguistic features and hidden states in each LSTM. Each multi-head attention in our CMA adopts
the same settings as those in AoA. Outputs of other projection layers in CMA share the same dimension of
1024. Final captions generated by CMA-DM are also limited to no more than 16 words.

\textbf{Training and Evaluation:} All trainable components in our framework are jointly optimized from
scratch with a trade-off coefficient. We train all of our models by first optimizing the cross entropy loss and then optimizing the
non-differentiable metrics with reinforcement learning. We use the Adam optimizer with batch size 50.
To avoid gradient explosion, we use the gradient clipping strategy \cite{sutskever2014sequence}, which truncates the gradients to a
range of [-0.1, 0.1]. As for the training process, we first train our models with cross entropy loss
for 30 epochs. We adopt label smoothing \cite{szegedy2016rethinking} and set the confidence to 0.2. We use an initial learning rate
of 2e-4 and anneal it by 0.8 every 3 epochs. We increase the scheduled sampling probability by 0.05
every 5 epochs until it reaches the probability of 0.5 \cite{bengio2015scheduled}. In this training stage, we set the trade-off
parameter $\lambda_{XE}$ to 0.2 to achieve the best performance. Then we optimize the CIDEr-D score with
SCST \cite{rennie2017self} for another 15 epochs. The initial learning rate is set to 2e-5 and annealed by
0.5 when the CIDEr-D score shows no improvement for 3 consecutive evaluating steps. We set the trade-off
parameter $\lambda_{RL}$ to 0.1 for sequence-level training. When training, we generate both draft
captions and final captions by using greedy decoding to reduce the time cost. When testing, we employ
beam search for better model performance and beam size is set to 3 for all models.

\section{Experimental Results}
\label{sec:exp_results}

\subsection{Comparing with Single-Pass Decoding Baselines}

We conduct three groups of experiments by employing each of the single-pass decoding based models mentioned in
Section \ref{exp_setup_drafting_models} as the \emph{Drafting Model}. As shown in Table. \ref{tab_baselines}, each group
contains the experimental results produced by four different methods: (1) \emph{method} is exactly the same as
the model in the original paper. (2) \emph{method$^{\sharp}$} and \emph{method$^{\ast}$} implement the baseline
models conforming to the structure setups of our \emph{Drafting Model}. More specifically, compared to the baseline
models, both \emph{Att2in$^{\sharp}$} and \emph{Up-Down$^{\sharp}$} contain an additional refining encoder,
which consists of a linear transformation layer followed by ReLU. As \emph{AoANet} fully conforms to the
structure of our \emph{Drafting Model}, \emph{AoANet$^{\ast}$} reports the results obtained from the publicly available
pre-trained AoANet. (3) \emph{method(Drafting)} is the \emph{Drafting Model} in our proposed framework, which performs
the first-pass decoding process and generates the draft captions. (4) \emph{method + CMA-DM} is our complete
two-pass decoding based model, which uses \emph{method$^{\sharp}$} or \emph{method$^{\ast}$} as the \emph{Drafting Model}.
In this section, we first compare the model performances in different groups to show the effect of our CMA-DM,
and then thoroughly analyze the characteristics of our proposed two-pass decoding framework by combining with
all experiments.

\textbf{Att2in Versus Att2in + CMA-DM:} In the first group of experiments, we apply our CMA-DM to the baseline
Att2in model. We report the results in Table. \ref{tab_baselines} Group 1. Compared to \emph{Att2in(Drafting)},
our \emph{Att2in + CMA-DM} leads to significant improvements across all metrics after different training stages,
especially increasing the CIDEr score from 111.7 to 118.7 with cross entropy loss and from 114.3 to 121.9 with
reinforcement learning.

\textbf{Up-Down Versus Up-Down + CMA-DM:} For the second group, we apply our CMA-DM to the popular Up-Down model.
As shown in Table. \ref{tab_baselines} Group 2, our \emph{Up-Down + CMA-DM} also achieves better performances
than its \emph{Drafting Model} \emph{Up-Down(Drafting)} across all metrics. \emph{Up-Down + CMA-DM} increases the
BLEU-4 / CIDEr scores from 35.9 / 113.0 to 37.5 / 115.1 with cross entropy loss and from 37.2 / 122.5 to 38.0 / 125.0
with reinforcement learning.

\textbf{AoANet Versus AoANet + CMA-DM:} We further apply our \emph{Deliberation Model} to the widely-used \emph{Drafting Model}
AoANet. The model performances reported in Table. \ref{tab_baselines} Group 3 show that our \emph{AoANet + CMA-DM}
outperforms \emph{AoANet(Drafting)} across most of the metrics, which substantially increases the BLEU-4 / CIDEr
scores from 36.8 / 116.2 to 38.7 / 120.2 with cross entropy loss.

\textbf{Framework Analysis: } We make a careful analysis of our proposed framework from three aspects: (1) By
comparing the performances between \emph{method(Drafting)} and \emph{method$^{\sharp}$} / \emph{method$^{\ast}$},
we discover that the former is often not as good as the latter. For example, considering the Up-Down model as
the baseline, the \emph{Drafting Model} \emph{Up-Down(Drafting)} achieves BLEU-4 / CIDEr scores of 35.9 / 113.0 with
cross entropy loss, while \emph{Up-Down$^{\sharp}$} obtains higher BLEU-4 / CIDEr scores of 36.2 / 113.7. This
is because the refining encoder of the \emph{Drafting Model} is also involved in the deliberation process. To ensure
the high performance of the \emph{Deliberation Model}, the refining encoder makes a compromise by sacrificing a little
performance of the \emph{Drafting Model}. (2) Comparison between \emph{method$^{\sharp}$} / \emph{method$^{\ast}$} and
our complete framework \emph{method + CMA-DM} demonstrates that the deliberation process is indeed effective at
polishing the draft captions, and even taking as input the slightly worse draft captions generated by \emph{method(Drafting)},
our CMA-DM can still achieve definite improvement over the \emph{method$^{\sharp}$} / \emph{method$^{\ast}$}. (3)
The better the performance of the baseline model, the less the improvement brought by the \emph{Deliberation Model}
with reinforcement learning. As reported in Table. \ref{tab_baselines}, \emph{Att2in$^{\sharp}$} achieves a CIDEr
score of 118.4 and \emph{Att2in + CMA-DM} shows an improvement of 3.0 percent on it. \emph{Up-Down$^{\sharp}$}
obtains a higher CIDEr score of 124.1 and \emph{Up-Down + CMA-DM} increases it by 0.7 percent. As for the
\emph{AoANet$^{\ast}$}, it achieves the performance of 128.9 CIDEr score and the deliberation process only provides
little improvement of 0.07 percent.

\begin{figure*}[!tp]
    \begin{center}
        \includegraphics[width=7.0in]{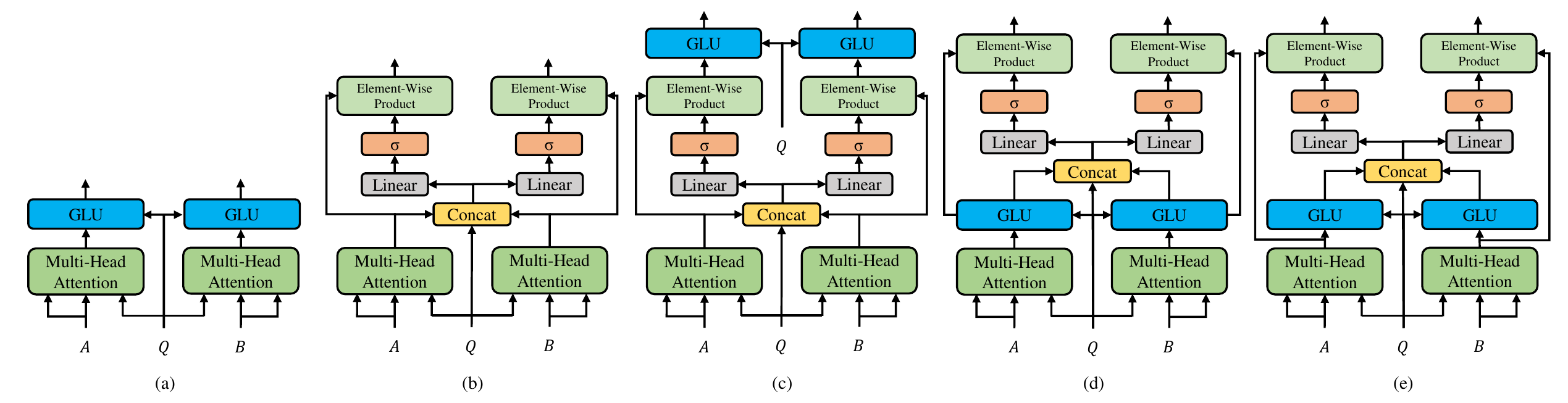}
    \end{center}
    \setlength{\abovecaptionskip}{0cm}
    \caption{We design five variants of the CMA module to figure out the best possible network structure for mutual correction.
    Note that in this group of comparisons, we do not add residual connections to any of these networks.}
    \label{fig_variants}
\end{figure*}
\begin{table*}[!tp]
    \begin{center}
        \caption{Ablation Performance of Our CMA Module. The results are reported after the cross entropy training stage.}
        \label{tab_ablation_study}
        \setlength{\tabcolsep}{1.3mm}{
        \begin{tabular}{ccccccccccccccc}
            \toprule
            \multicolumn{1}{c}{} &{} &\multicolumn{2}{c}{Attention on} &\multicolumn{2}{c}{Residual Connection on} &\multicolumn{9}{c}{Metric} \\
            \cmidrule(lr){3-4}
            \cmidrule(lr){5-6}
            \cmidrule(lr){7-15}
            \multicolumn{1}{c}{Number} &{Attention Module} &{Visual Side} &{Textual Side} &{Visual Side} &{Textual Side} &{B@1} &{B@2} &{B@3} &{B@4} &{M} &{R} &{C} &{S} &{W} \\
            \midrule
            \midrule
            \multicolumn{1}{c|}{1} &\multicolumn{1}{c|}{}
            &\multicolumn{1}{c}{\Checkmark} &\multicolumn{1}{c|}{\XSolidBrush} &\multicolumn{1}{c}{\XSolidBrush} &\multicolumn{1}{c|}{\XSolidBrush}
            &\multicolumn{1}{c}{77.0} &{61.2} &{47.4} &{36.8} &{\textbf{28.0}} &{57.0} &{116.4} &{\textbf{20.9}} &{\textbf{25.8}} \\
            \multicolumn{1}{c|}{2} &\multicolumn{1}{c|}{Multi-Head Attention}
            &\multicolumn{1}{c}{\XSolidBrush} &\multicolumn{1}{c|}{\Checkmark} &\multicolumn{1}{c}{\XSolidBrush} &\multicolumn{1}{c|}{\XSolidBrush}
            &\multicolumn{1}{c}{75.1} &{59.1} &{45.6} &{35.2} &{27.2} &{56.0} &{110.2} &{20.0} &{24.6} \\
            \multicolumn{1}{c|}{3} &\multicolumn{1}{c|}{}
            &\multicolumn{1}{c}{\Checkmark} &\multicolumn{1}{c|}{\Checkmark} &\multicolumn{1}{c}{\XSolidBrush} &\multicolumn{1}{c|}{\XSolidBrush}
            &\multicolumn{1}{c}{\textbf{77.2}} &{\textbf{61.5}} &{\textbf{47.8}} &{\textbf{37.0}} &{\textbf{28.0}} &{\textbf{57.2}} &{\textbf{116.8}} &{\textbf{20.9}} &{\textbf{25.8}} \\
            \midrule
            \multicolumn{1}{c|}{4} &\multicolumn{1}{c|}{CMA-(a)}
            &\multicolumn{1}{c}{\Checkmark} &\multicolumn{1}{c|}{\Checkmark} &\multicolumn{1}{c}{\XSolidBrush} &\multicolumn{1}{c|}{\XSolidBrush}
            &\multicolumn{1}{c}{76.9} &{61.2} &{47.7} &{37.0} &{28.0} &{57.1} &{116.3} &{20.9} &{25.8} \\
            \multicolumn{1}{c|}{5} &\multicolumn{1}{c|}{CMA-(b)}
            &\multicolumn{1}{c}{\Checkmark} &\multicolumn{1}{c|}{\Checkmark} &\multicolumn{1}{c}{\XSolidBrush} &\multicolumn{1}{c|}{\XSolidBrush}
            &\multicolumn{1}{c}{77.8} &{62.5} &{48.5} &{37.3} &{27.5} &{57.5} &{117.0} &{20.8} &{25.5} \\
            \multicolumn{1}{c|}{6} &\multicolumn{1}{c|}{CMA-(c)}
            &\multicolumn{1}{c}{\Checkmark} &\multicolumn{1}{c|}{\Checkmark} &\multicolumn{1}{c}{\XSolidBrush} &\multicolumn{1}{c|}{\XSolidBrush}
            &\multicolumn{1}{c}{76.9} &{61.2} &{47.6} &{37.0} &{28.0} &{57.0} &{116.5} &{20.8} &{25.7} \\
            \multicolumn{1}{c|}{7} &\multicolumn{1}{c|}{CMA-(d)}
            &\multicolumn{1}{c}{\Checkmark} &\multicolumn{1}{c|}{\Checkmark} &\multicolumn{1}{c}{\XSolidBrush} &\multicolumn{1}{c|}{\XSolidBrush}
            &\multicolumn{1}{c}{\textbf{78.2}} &{\textbf{62.9}} &{\textbf{48.9}} &{\textbf{37.8}} &{27.8} &{\textbf{57.7}} &{\textbf{118.2}} &{21.0} &{26.0} \\
            \multicolumn{1}{c|}{8} &\multicolumn{1}{c|}{CMA-(e)}
            &\multicolumn{1}{c}{\Checkmark} &\multicolumn{1}{c|}{\Checkmark} &\multicolumn{1}{c}{\XSolidBrush} &\multicolumn{1}{c|}{\XSolidBrush}
            &\multicolumn{1}{c}{76.8} &{61.0} &{47.4} &{36.8} &{\textbf{28.1}} &{57.0} &{116.5} &{\textbf{21.1}} &{\textbf{26.0}} \\
            \midrule
            \multicolumn{1}{c|}{9} &\multicolumn{1}{c|}{}
            &\multicolumn{1}{c}{\Checkmark} &\multicolumn{1}{c|}{\Checkmark} &\multicolumn{1}{c}{\Checkmark} &\multicolumn{1}{c|}{\XSolidBrush}
            &\multicolumn{1}{c}{\textbf{78.2}} &{\textbf{63.1}} &{\textbf{49.4}} &{\textbf{38.3}} &{\textbf{28.1}} &{\textbf{57.9}} &{\textbf{119.7}} &{\textbf{21.2}} &{\textbf{26.1}} \\
            \multicolumn{1}{c|}{10} &\multicolumn{1}{c|}{CMA-(d)}
            &\multicolumn{1}{c}{\Checkmark} &\multicolumn{1}{c|}{\Checkmark} &\multicolumn{1}{c}{\XSolidBrush} &\multicolumn{1}{c|}{\Checkmark}
            &\multicolumn{1}{c}{77.0} &{60.9} &{47.2} &{36.6} &{27.7} &{56.9} &{115.6} &{20.8} &{25.6} \\
            \multicolumn{1}{c|}{11} &\multicolumn{1}{c|}{}
            &\multicolumn{1}{c}{\Checkmark} &\multicolumn{1}{c|}{\Checkmark} &\multicolumn{1}{c}{\Checkmark} &\multicolumn{1}{c|}{\Checkmark}
            &\multicolumn{1}{c}{77.4} &{61.5} &{47.9} &{37.1} &{27.8} &{57.3} &{116.6} &{20.9} &{25.8} \\
            \bottomrule
        \end{tabular}}
    \end{center}
\end{table*}

\subsection{Ablation Study on CMA}

To clearly explain the performance improvement brought by each component in our CMA, we conduct ablation study
by employing AoANet as the \emph{Drafting Model} and the standard top-down architecture \cite{anderson2018bottom} as the deliberation decoder.
We evaluate different methods by only changing the structure of the attention module. We design five variants
of the CMA module, as illustrated in Fig. \ref{fig_variants}, and determine the best possible network structure by comparing
their performances. We report the results after optimizing cross entropy loss with the trade-off parameter
$\lambda_{XE} = 1.0$ in Table. \ref{tab_ablation_study}.

We first make experiments (1 - 3) based on the conventional multi-head attention mechanism. Experiment 1 adopts
the original structure of the top-down decoder, which contains a visual attention mechanism like the decoders
of most single-pass decoding based models to capture visual context features for each time step. Experiment 2
applies the multi-head attention function only to the features of draft captions to decide which parts of the
linguistic features are the most relevant to the word currently to be generated. Experiment 3 simultaneously
uses visual attention mechanism and textual attention mechanism to selectively focus on certain parts of both
visual and linguistic features. The experimental results show that the model only equipped with textual attention
mechanism in Experiment 2 reaches poor performances and gets a CIDEr score of 110.2. As expected, however, the
performance of the model integrated with both visual and textual attention mechanisms in Experiment 3 performs
slightly better than those reported in Experiment 1 and 2.

Considering the significance and synergism of both visual and textual attention mechanisms, we design five variants
of the CMA module based on this parallel attention structure, as illustrated in Fig. \ref{fig_variants}. These modules perform
mutual correction between visual and linguistic context features in different ways, which mainly depends on the
sigmoid activation function. The comparison results are reported in the experiments (4 - 8). We weigh up the
performances of different CMA modules and decide to adopt \emph{CMA-(d)} as the main structure of our CMA, which
achieves the best performances on most metrics.

We further test the performances of residual connections. We conduct experiments (9 - 11) by applying the residual
connection to the visual side, textual side, or both sides of \emph{CMA-(d)}. Fig. \ref{fig_method_attentions} (b)
illustrates how the residual connection is added to \emph{CMA-(d)} on its visual side, and it is a symmetrical
structure for the residual connection on the textual side. The experimental results show that the residual connection on the visual side brings
great improvements across all metrics, while the residual connection on the textual side leads to performance
degradation.

\subsection{Comparison of Different Complexity Refining Encoders}

To gain insight into the influences of the refining encoder with different complexity, we conduct comparative
experiments with the same setups, except the network structure of the refining encoder. This group of experiments
are also based on the AoANet, and we use the \emph{CMA-(d)} with a residual connection applied to the visual
features as the deliberation decoder. We evaluate the performances of three different refining encoders: (1) The
first kind of refining encoder is composed of a single linear transformation layer with ReLU as its activation
function. (2) The second one consists of 3 layers of the AoA module. (3) The last one adopts the same structure
as that in the original AoANet, which contains 6 layers of the AoA module. The comparison results are reported in
Table. \ref{tab_refining_encoders} after cross entropy training with $\lambda_{XE} = 1.0$. We find that the more
complex the refining encoder, the higher the performance of the two models.

\begin{table}[!tp]
    \begin{center}
        \caption{Performance Comparisons of Different Complexity Refining Encoders with AoANet \cite{huang2019attention}
        as Our Drafting Model.}
        \label{tab_refining_encoders}
        \setlength{\tabcolsep}{1.3mm}{
        \begin{tabular}{cccccc}
            \toprule
            \multicolumn{1}{c}{Refining Encoder} &{Method} &{B@4} &{R} &{C} &{S} \\
            \midrule
            \midrule
            \multirow{2}{*}{FC + ReLU}
            &\multicolumn{1}{|l|}{AoANet(Drafting)} &{36.2} &{56.5} &{114.1} &{20.6} \\
            &\multicolumn{1}{|l|}{AoANet + CMA-DM} &{37.7} &{57.6} &{117.2} &{20.9} \\
            \midrule
            \multirow{2}{*}{3 $\times$ AoA}
            &\multicolumn{1}{|l|}{AoANet(Drafting)} &{36.6} &{56.8} &{116.7} &{20.8} \\
            &\multicolumn{1}{|l|}{AoANet + CMA-DM} &{38.0} &{57.6} &{117.9} &{21.0} \\
            \midrule
            \multirow{2}{*}{6 $\times$ AoA}
            &\multicolumn{1}{|l|}{AoANet(Drafting)} &{36.9} &{56.9} &{117.2} &{20.9} \\
            &\multicolumn{1}{|l|}{AoANet + CMA-DM} &\textbf{38.3} &\textbf{57.9} &\textbf{119.7} &\textbf{21.2} \\
            \bottomrule
        \end{tabular}}
    \end{center}
\end{table}

\subsection{Selection of Trade-Off Parameters}

We evaluate the two-pass decoding framework with 7 different trade-off coefficients in both training stages, as shown
in Table \ref{tab_trade_off_parameters}. All experiments are conducted based on the AoANet and employ the complete
structure described in Section \ref{sec:method}. We report the results of both drafting models and deliberation models.
For the training stage with cross entropy loss, as the value of $\lambda_{XE}$ decreases, the performance of the
\emph{Drafting Model} becomes increasingly better. The \emph{Drafting Model} achieves the best performance when $\lambda_{XE}$ = 0.2.
The deliberation models follow the same rule in the cross entropy training stage. When testing models with different
coefficients under SCST, we employ the same model pre-trained after cross entropy loss with $\lambda_{XE}$ = 0.2 as initialization.
We find that the performances of drafting models hardly changes with variation of $\lambda_{RL}$, while the
\emph{Deliberation Model} still has obvious improvement as $\lambda_{RL}$ decreases.

\begin{table}[!tp]
    \begin{center}
        \caption{Performance Comparisons of Our Framework with Different Trade-Off Coefficient. Best Results of the Drafting Model
        and the Deliberation Model are Highlighted in Red and Blue, Respectively.}
        \label{tab_trade_off_parameters}
        \setlength{\tabcolsep}{1.2mm}{
        \begin{tabular}{cccccccc}
            \toprule
            \multicolumn{1}{c}{Trade-Off} &{} &\multicolumn{3}{c}{Cross Entropy Loss} &\multicolumn{3}{c}{SCST} \\
            \cmidrule(lr){3-5}
            \cmidrule(lr){6-8}
            \multicolumn{1}{c}{Coefficient} &{Method} &{B@4} &{R} &{C} &{B@4} &{R} &{C} \\
            \midrule
            \midrule
            \multicolumn{1}{c}{\multirow{2}{*}{$\lambda$ = 10.0}}
            &\multicolumn{1}{|c|}{Drafting} &{35.2} &{56.1} &{112.3}
            &\multicolumn{1}{|c}{38.7} &{58.6} &{127.9} \\
            &\multicolumn{1}{|c|}{Deliberation} &{37.1} &{57.0} &{115.2}
            &\multicolumn{1}{|c}{38.2} &{58.3} &{125.6} \\
            \midrule
            \multicolumn{1}{c}{\multirow{2}{*}{$\lambda$ = 5.0}}
            &\multicolumn{1}{|c|}{Drafting} &{35.3} &{56.2} &{113.4}
            &\multicolumn{1}{|c}{38.7} &{58.6} &{128.1} \\
            &\multicolumn{1}{|c|}{Deliberation} &{37.6} &{57.4} &{116.9}
            &\multicolumn{1}{|c}{38.5} &{58.4} &{126.4} \\
            \midrule
            \multicolumn{1}{c}{\multirow{2}{*}{$\lambda$ = 2.0}}
            &\multicolumn{1}{|c|}{Drafting} &{35.8} &{56.5} &{115.1}
            &\multicolumn{1}{|c}{38.9} &{58.7} &{127.9} \\
            &\multicolumn{1}{|c|}{Deliberation} &{38.0} &{57.6} &{118.0}
            &\multicolumn{1}{|c}{38.5} &{58.4} &{126.5} \\
            \midrule
            \multicolumn{1}{c}{\multirow{2}{*}{$\lambda$ = 1.0}}
            &\multicolumn{1}{|c|}{Drafting} &{36.3} &{56.8} &{115.6}
            &\multicolumn{1}{|c}{38.9} &{58.7} &{128.3} \\
            &\multicolumn{1}{|c|}{Deliberation} &{38.3} &{57.9} &{119.7}
            &\multicolumn{1}{|c}{38.6} &{58.5} &{127.5} \\
            \midrule
            \multicolumn{1}{c}{\multirow{2}{*}{$\lambda$ = 0.5}}
            &\multicolumn{1}{|c|}{Drafting} &{36.8} &{56.7} &{116.3}
            &\multicolumn{1}{|c}{38.8} &{58.7} &{127.8} \\
            &\multicolumn{1}{|c|}{Deliberation} &{38.0} &{57.7} &{118.5}
            &\multicolumn{1}{|c}{38.7} &{58.5} &{127.8} \\
            \midrule
            \multicolumn{1}{c}{\multirow{2}{*}{$\lambda$ = 0.2}}
            &\multicolumn{1}{|c|}{Drafting} &\textbf{\textcolor{red}{37.0}} &\textbf{\textcolor{red}{57.1}}
            &\textbf{\textcolor{red}{116.7}}
            &\multicolumn{1}{|c}{38.7} &{58.5} &{127.9} \\
            &\multicolumn{1}{|c|}{Deliberation} &\textbf{\textcolor{blue}{38.7}} &\textbf{\textcolor{blue}{58.2}}
            &\textbf{\textcolor{blue}{120.2}}
            &\multicolumn{1}{|c}{38.9} &{58.7} &{128.4} \\
            \midrule
            \multicolumn{1}{c}{\multirow{2}{*}{$\lambda$ = 0.1}}
            &\multicolumn{1}{|c|}{Drafting} &{36.8} &{57.0} &{116.3}
            &\multicolumn{1}{|c}{\textbf{\textcolor{red}{39.0}}}  &\textbf{\textcolor{red}{58.8}} &\textbf{\textcolor{red}{128.3}} \\
            &\multicolumn{1}{|c|}{Deliberation} &{38.4} &{58.1} &{119.7}
            &\multicolumn{1}{|c}{\textbf{\textcolor{blue}{39.2}}} &\textbf{\textcolor{blue}{58.9}} &\textbf{\textcolor{blue}{129.0}} \\
            \bottomrule
        \end{tabular}}
    \end{center}
\end{table}

\subsection{Comparing With State-of-The-Art Methods}

\begin{table*}[tp]
    \begin{center}
        \caption{Performance of Our Method and Other State-of-the-Art Methods on MS COCO ``Karpathy'' Test Split.}
        \label{table_sota}
        \setlength{\tabcolsep}{2.0mm}{
        \begin{tabular}{lcccccccccccc}
            \toprule
            \multicolumn{1}{c}{} &\multicolumn{6}{c}{Cross Entropy Loss} &\multicolumn{6}{c}{Reinforcement Learning} \\
            \cmidrule(lr){2-7}
            \cmidrule(lr){8-13}
            \multicolumn{1}{c}{Model} &\multicolumn{1}{c}{B@1} &{B@4} &{M} &{R} &{C} &\multicolumn{1}{c|}{S}
            &\multicolumn{1}{c}{B@1} &{B@4} &{M} &{R} &{C} &{S} \\
            \midrule
            \midrule
            \multicolumn{1}{l}{NIC \cite{vinyals2016show}} &{-} &{29.6} &{25.2} &{52.6} &{94.0} &\multicolumn{1}{c|}{-}
            &{-} &{31.9} &{25.5} &{54.3} &{106.3} &{-} \\
            \multicolumn{1}{l}{Att2in \cite{rennie2017self}} &{-} &{31.3} &{26.0} &{54.3} &{101.3} &\multicolumn{1}{c|}{-}
            &{-} &{33.3} &{26.3} &{55.3} &{111.4} &{-} \\
            \multicolumn{1}{l}{Up-Down \cite{anderson2018bottom}} &{77.2} &{36.2} &{27.0} &{56.4} &{113.5} &\multicolumn{1}{c|}{20.3}
            &{79.8} &{36.3} &{27.7} &{56.9} &{120.1} &{21.4} \\
            \multicolumn{1}{l}{DA \cite{gao2019deliberate}} &{75.8} &{35.7} &{27.4} &{56.2} &{111.9} &\multicolumn{1}{c|}{20.5}
            &{79.9} &{37.5} &{28.5} &{58.2} &{125.6} &{22.3} \\
            \multicolumn{1}{l}{AoANet \cite{huang2019attention}} &{77.4} &{37.2} &{28.4} &{57.5} &{119.8} &\multicolumn{1}{c|}{21.3}
            &{80.2} &{38.9} &\textbf{29.2} &{58.8} &\textbf{129.8} &{22.4} \\
            \multicolumn{1}{l}{AoANet$^{\ast}$} &{77.3} &{36.9} &\textbf{28.5} &{57.3} &{118.4} &\multicolumn{1}{c|}{21.7}
            &{80.5} &{39.1} &{29.0} &\textbf{58.9} &{128.9} &\textbf{22.7} \\
            \multicolumn{1}{l}{Up-Down + MN \cite{sammani2019look}} &{76.9} &{36.1} &{-} &{56.4} &{112.3} &\multicolumn{1}{c|}{20.3}
            &{-} &{-} &{-} &{-} &{-} &{-} \\
            \multicolumn{1}{l}{Up-Down + RD \cite{guo2019show}} &{-} &{36.7} &{27.8} &{56.8} &{114.5} &\multicolumn{1}{c|}{20.8}
            &{80.0} &{37.8} &{28.2} &{57.9} &{125.3} &{21.7} \\
            \multicolumn{1}{l}{GCN-LSTM$_{spa}$ + RD \cite{guo2019show}} &{-} &{37.0} &{27.9} &{57.1} &{116.1} &\multicolumn{1}{c|}{21.0}
            &{80.5} &{38.6} &{28.7} &{58.7} &{128.3} &{22.3} \\
            \multicolumn{1}{l}{AoANet + ETN \cite{sammani2020show}} &{77.9} &{38.0} &{-} &{57.7} &{120.0} &\multicolumn{1}{c|}{21.2}
            &\textbf{80.6} &\textbf{39.2} &{-} &\textbf{58.9} &{128.9} &{22.6} \\
            \midrule
            \multicolumn{1}{l}{Up-Down + CMA-DM} &{78.0} &{37.5} &{27.8} &{57.2} &{115.1} &\multicolumn{1}{c|}{20.9}
            &{80.1} &{38.0} &{28.3} &{58.2} &{125.0} &{21.7} \\
            \multicolumn{1}{l}{AoANet + CMA-DM} &\textbf{78.6} &\textbf{38.7} &{28.4} &\textbf{58.2} &\textbf{120.2} &\multicolumn{1}{c|}{\textbf{26.2}}
            &\textbf{80.6} &\textbf{39.2} &{29.0} &\textbf{58.9} &{129.0} &{22.6} \\
            \bottomrule
        \end{tabular}}
    \end{center}
\end{table*}
\begin{table*}[tp]
    \begin{center}
        \caption{Leaderboard of Various Published Models on the Online COCO Test Server.}
        \label{table_online_test}
        \setlength{\tabcolsep}{2.0mm}{
        \begin{tabular}{lcccccccccccccc}
            \toprule
            \multicolumn{1}{l}{Model} &\multicolumn{2}{c}{BLEU-1} &\multicolumn{2}{c}{BLEU-2} &\multicolumn{2}{c}{BLEU-3} &\multicolumn{2}{c}{BLEU-4}
            &\multicolumn{2}{c}{METEOR} &\multicolumn{2}{c}{ROUGE\_L} &\multicolumn{2}{c}{CIDEr} \\
            \cmidrule(lr){2-3}
            \cmidrule(lr){4-5}
            \cmidrule(lr){6-7}
            \cmidrule(lr){8-9}
            \cmidrule(lr){10-11}
            \cmidrule(lr){12-13}
            \cmidrule(lr){14-15}
            \multicolumn{1}{l}{Metric} &\multicolumn{1}{c}{c5} &\multicolumn{1}{c}{c40} &\multicolumn{1}{c}{c5} &\multicolumn{1}{c}{c40}
            &\multicolumn{1}{c}{c5} &\multicolumn{1}{c}{c40} &\multicolumn{1}{c}{c5} &\multicolumn{1}{c}{c40} &\multicolumn{1}{c}{c5}
            &\multicolumn{1}{c}{c40} &\multicolumn{1}{c}{c5} &\multicolumn{1}{c}{c40} &\multicolumn{1}{c}{c5} &\multicolumn{1}{c}{c40} \\
            \midrule
            \midrule
            \multicolumn{1}{l}{SCST \cite{rennie2017self}} &{78.1} &{93.7} &{61.9} &{86.0} &{47.0} &{75.9} &{35.2} &{64.5} &{27.0} &{35.5} &{56.3} &{70.7} &{114.7} &{116.0} \\
            \multicolumn{1}{l}{Up-Down \cite{anderson2018bottom}} &{80.2} &{95.2} &{64.1} &{88.8} &{49.1} &{79.4} &{36.9} &{68.5} &{27.6} &{36.7} &{57.1} &{72.4} &{117.9} &{120.5} \\
            \multicolumn{1}{l}{AoANet \cite{huang2019attention}} &{81.0} &{95.0} &{65.8} &{89.6} &{51.4} &{81.3} &{39.4} &{71.2} &{29.1} &{38.5} &{58.9} &{74.5} &{126.9} &{129.6} \\
            \multicolumn{1}{l}{DA \cite{gao2019deliberate}} &{79.4} &{94.4} &{63.5} &{88.0} &{48.7} &{78.4} &{36.8} &{67.4} &{28.2} &{37.0} &{57.7} &{72.2} &{120.5} &{122.0} \\
            \multicolumn{1}{l}{Up-Down + RD \cite{guo2019show}} &{79.9} &{94.4} &{64.4} &{88.3} &{49.8} &{79.3} &{37.9} &{68.8} &{28.4} &{37.5} &{58.1} &{73.1} &{122.2} &{124.1} \\
            \multicolumn{1}{l}{GCN-LSTM$_{spa}$ + RD \cite{guo2019show}} &{80.0} &{94.7} &{64.5} &{88.6} &{49.8} &{79.5} &{37.8} &{68.8} &{28.4} &{37.6} &{58.0} &{73.2} &{123.8} &{126.4} \\
            \midrule
            \multicolumn{1}{l}{AoANet + CMA-DM} &{79.9} &{94.6} &{64.7} &{88.8} &{50.2} &{80.0} &{38.3} &{69.5} &{28.3} &{37.5} &{58.2} &{73.2} &{122.9} &{125.2} \\
            \bottomrule
        \end{tabular}}
    \end{center}
\end{table*}

We compare our models, including \emph{Up-Down + CMA-DM} and \emph{AoANet + CMA-DM}, with current state-of-the-art on
the MS COCO ``Karpathy'' test split, especially other two-pass decoding based methods. These models include (1) NIC \cite{vinyals2016show}, which
uses a vanilla CNN-LSTM encoder-decoder framework without attention mechanism; (2) Att2in \cite{rennie2017self}, which applies visual attention
mechanism and is optimized with non-differentiable metrics; (3) Up-Down \cite{anderson2018bottom}, which first extracts object-level features using
Faster R-CNN and then adopts a top-down decoder to decode the features; (4) DA \cite{gao2019deliberate}, which refines the context vectors by a
deliberate residual-based attention layer; (5) AoANet \cite{huang2019attention}, which uses an attention gate to retain the expect useful knowledge
from the attended results; (6) MN \cite{sammani2019look}, which modifies the existing captions from a given framework; (7) RD \cite{guo2019show},
which polishes the captions generated from a base model by using a ruminant decoder; (8) ETN \cite{sammani2020show}, which edits existing captions
with iterative adaptive refinement.

As shown in Table. \ref{table_sota}, our \emph{Up-Down + CMA-DM} dramatically outperforms its baseline model, and our
\emph{AoANet + CMA-DM} achieves better performances than \emph{AoANet($^{\ast}$)} on most of the metrics. Both results
indicate that our framework can really refine the draft captions generated by single-pass decoding based models. With the
same baseline model \emph{Up-Down}, our CMA-DM is superior to both MN and RD. Considering AoANet as the baseline, our
approach achieves competitive performance compared to the current state-of-the-art two-pass decoding based model ETN.
Furthermore, we report the model performances on the online COCO test server in Table. \ref{table_online_test}. The performance of our single
model \emph{AoANet + CMA-DM} is slightly lower than that of its baseline, which is an ensemble of 4 models.




\subsection{Qualitative Analysis}

\begin{figure*}[!tp]
    \begin{center}
        \includegraphics[width=5.5in]{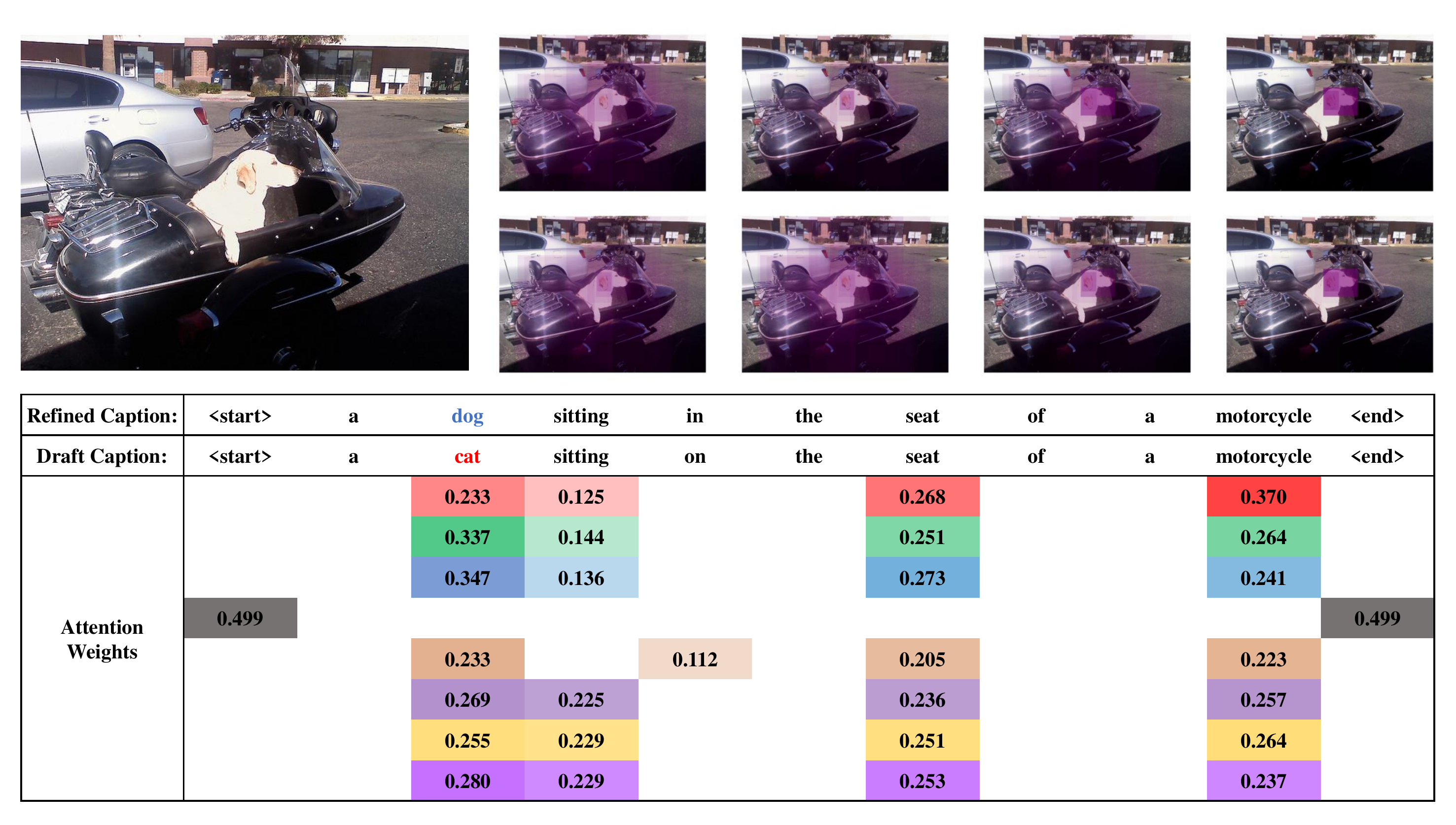}
    \end{center}
    \setlength{\abovecaptionskip}{0cm}
    \caption{Visualization of both visual and textual attention matrices in our CMA of \emph{AoANet + CMA-DM} model.
    For clarity, we show the attention maps of all 8 heads at the time step of generating the word, ``dog''. For
    textual attention visualization, we only report the attention results with weights higher than 0.1.}
    \label{fig_dog}
\end{figure*}

\textbf{Attention Visualization:} To qualitatively show the effect of our CMA-DM, we visualize both visual
and textual attention weights in Fig. \ref{fig_dog}. As our CMA extends the multi-head attention mechanism,
we only illustrate the attention maps at the time step of mistake correction for clarity. As shown in Fig.
\ref{fig_dog}, the draft caption generated in the first-pass decoding process contains an obvious mistake,
which is because the \emph{Drafting Model} mistakenly considers the dog on the motorcycle as a ``cat''. When
our CMA-DM attempts to predict the correct word ``dog'' in the polishing process, the CMA pays more attention
to the region of the dog in the image and assigns high weights to the related words in the draft caption,
including ``cat'', ``seat'' and ``motorcycle''. In the correction process, the visual context provides more
original and reliable information related to the animal and filters out the features related to ``cat'' from
the linguistic context. As a result, our CMA-DM finally outputs ``dog'' for the next word instead of the ``cat''
and refines the draft caption to a better descriptive sentence.

\begin{figure*}[!tp]
    \begin{center}
        \includegraphics[width=5.5in]{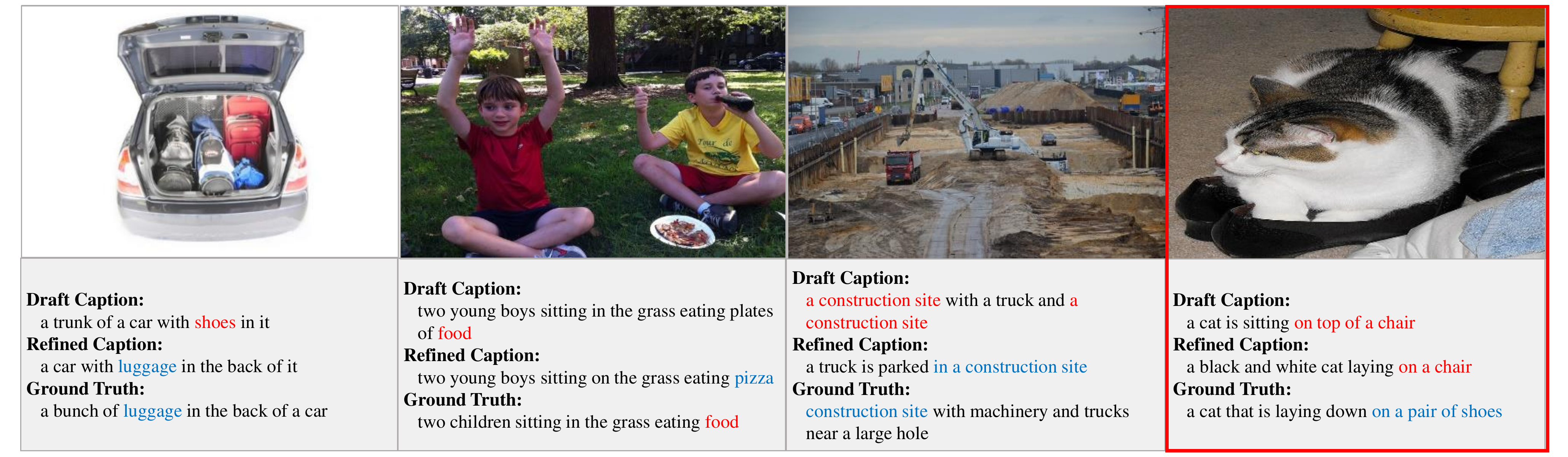}
    \end{center}
    \setlength{\abovecaptionskip}{0cm}
    \caption{Examples of image captions generated by \emph{AoANet + CMA-DM} model along with the corresponding
    ground truths. Mistakes or inappropriate words and phrases are highlighted in red. Red box indicates failure
    case.}
    \label{fig_captions}
\end{figure*}

\textbf{Generated Captions:} To further demonstrate the improvement of our CMA-DM on refining draft captions,
we show some qualitative examples generated by our two-pass decoding based framework in Fig. \ref{fig_captions}.
For each example, we show the draft caption generated by the \emph{Drafting Model} and the refined caption generated
by our CMA-DM, along with the ground-truth caption for comparison.
In the first example, our CMA-DM corrects the mistake, ``shoes'', to a more accurate word ``luggage''.
In the second example, our CMA-DM modifies the details in the draft captions (i.e., ``food'' to ``pizza'') and
even makes more detailed expression than the ground truth. As our CMA-DM can make global planning based on the draft
captions, we can produce descriptions with higher degree of semantic coherency and fluency for given images.
The third example makes a specific explanation of this. The phrase, ``a construction site'', appears twice
in the draft caption, while our refined caption does not suffer this problem. We also provide a failure case
in fourth example, where both draft and refined captions miss the salient object, ``a pair of shoes'',
which possibly results from the occlusion of the cat.


\textbf{Human Evaluation:}
For human evaluation, we randomly select 300 images from MS COCO ``Karpathy'' validation set and provide the pairwise
captions generated by the \emph{Drafting Model} and our CMA-DM to 5 different evaluators along with the image. We perform the
evaluation based on Up-Down and AoANet. We report relevance and coherence for both captions in Table. \ref{tab_human_evaluation}.
Relevance measures the correlation degree between the generated caption and the image content, while coherence indicates
the fluency and semantic consistency of the captions. The evaluators are required to choose which caption is better or
they achieve the competitive performance. We consider a caption ``Is Better'' only if it receives more than 2 positive
replies. Evaluation results reported in Table. \ref{tab_human_evaluation} show that our CMA-DM achieves significant
improvement over the drafting models in terms of both relevance and coherence.

\begin{table}
    \begin{center}
        \caption{Results of Human Evaluation.}
        \label{tab_human_evaluation}
        \setlength{\tabcolsep}{1.2mm}{
        \begin{tabular}{lcc}
            \toprule
            \multicolumn{1}{l}{Model} &{Relevance} &{Coherence} \\
            \midrule
            \midrule
            \multicolumn{1}{l}{Up-Down(Drafting)} &{9.3} &{8.7} \\
            \multicolumn{1}{l}{Up-Down + CMA-DM} &{14.0} &{18.7} \\
            \midrule
            \multicolumn{1}{l}{AoANet(Drafting)} &{7.3} &{6.7} \\
            \multicolumn{1}{l}{AoANet + CMA-DM} &{9.3} &{12.3} \\
            \bottomrule
        \end{tabular}}
    \end{center}
\end{table}

\section{Conclusion}
\label{sec:conclusion}
In this paper, we explore a universal two-pass decoding based framework for
image captioning to overcome the limitations of the single-pass decoding based
models. To enhance the semantic expression of visual features and filter out
the error information in the draft captions, we integrate a well-designed Cross
Modification Attention (CMA) module with the docder the \emph{Deliberation Model}
to perform mutual correction between the features of input images and corresponding
draft captions. Experiments on MS COCO dataset demonstrate the advantages and
improvements of our CMA and the proposed framework.


%





\ifCLASSOPTIONcaptionsoff
  \newpage
\fi



%



\bibliographystyle{IEEEtran}
\bibliography{reference}

%
\vspace{-13 mm}
\begin{IEEEbiography}[{\includegraphics[width=1in,height=1.25in,clip,keepaspectratio]{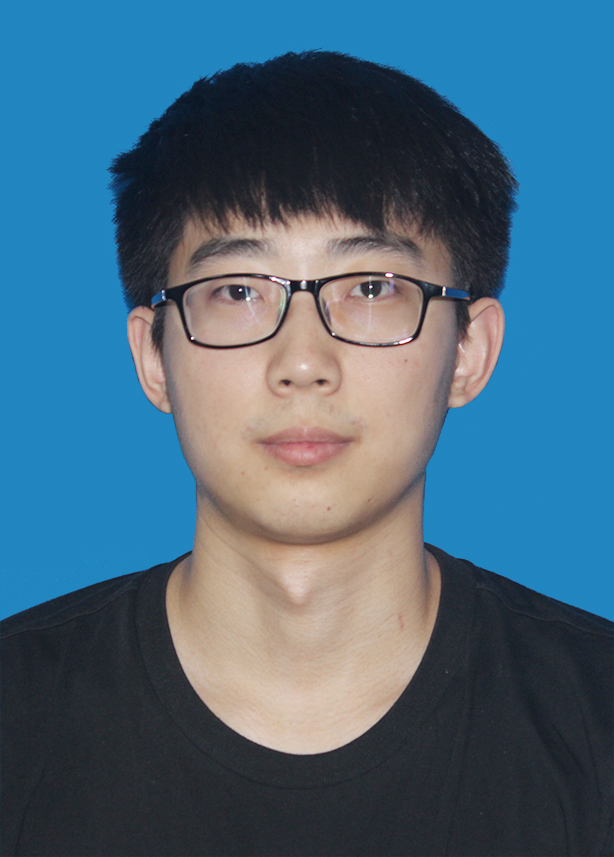}}]{Zheng Lian}
Zheng Lian received the B.E. degree from Xidian University, Shaanxi, China, in 2017. He is currently working
toward the Ph.D. degree with the Science \& Technology on Integrated Information System Laboratory, Institute of
Software Chinese Academy of Sciences, Beijing, China. He is now majoring in deep learning and image
captioning.
\end{IEEEbiography}
\vspace{-13 mm}
\begin{IEEEbiography}[{\includegraphics[width=1in,height=1.25in,clip,keepaspectratio]{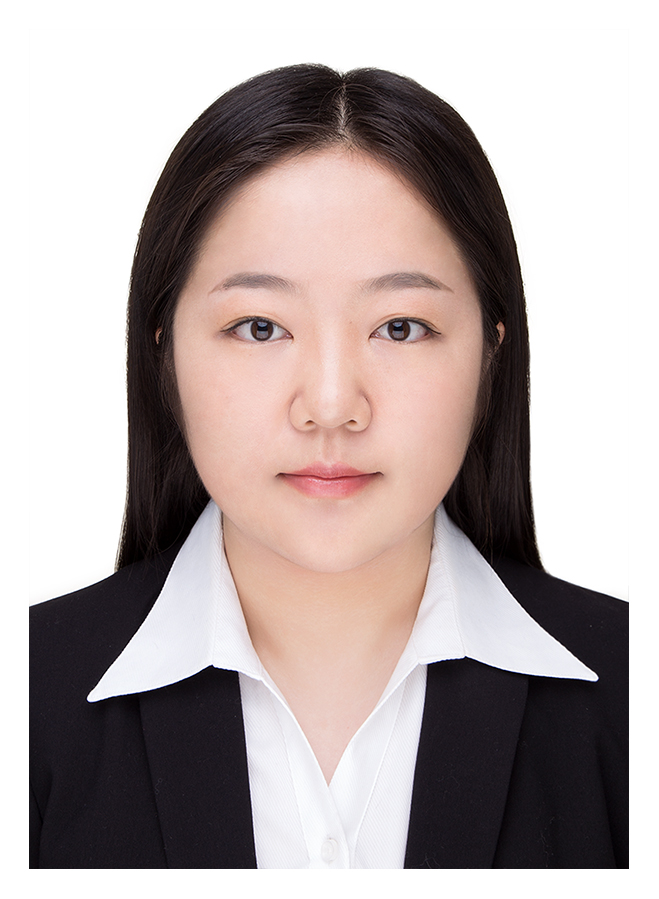}}]{Yanan Zhang}
Yanan Zhang received the B.E. degree from Xiamen University, Fujian, China, in 2018. She is currently working
toward the Ph.D. degree with the Science \& Technology on Integrated Information System Laboratory, Institute of
Software Chinese Academy of Sciences, Beijing, China. She is now majoring in deep learning and object
detection.
\end{IEEEbiography}
\vspace{-13 mm}
\begin{IEEEbiography}[{\includegraphics[width=1in,height=1.25in,clip,keepaspectratio]{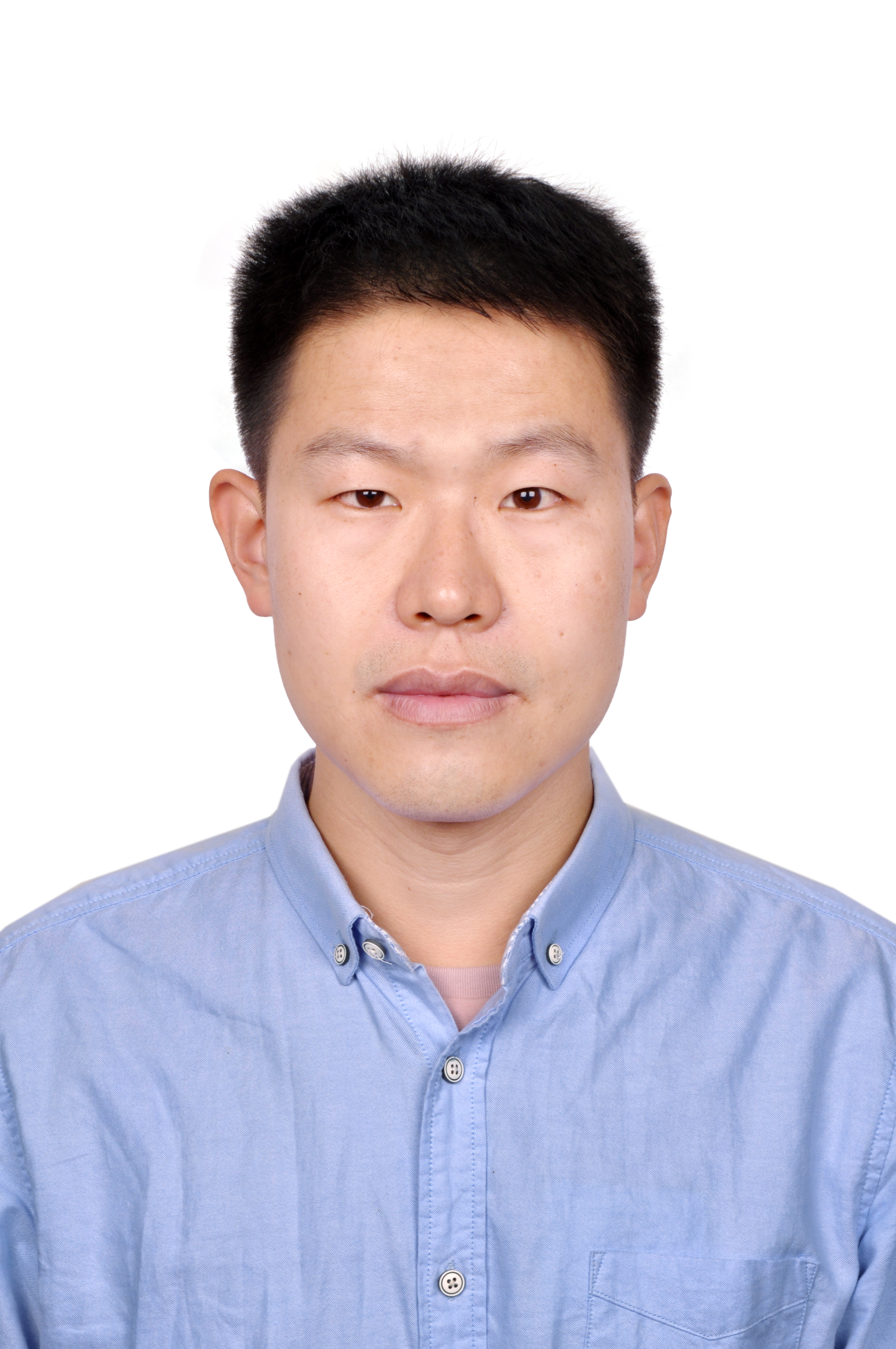}}]{Haichang Li}
Haichang Li received the B.S. degree in mathematics from Shandong University, Shandong, China, in 2003, and
the M.S. degree and the Ph.D. degree from the Institute of Automation, Chinese Academy of Sciences, Beijing,
China in 2011 and 2016, respectively. He is currently an Associated Professor with Science \& Technology on
Integrated Information System Laboratory, Institute of Software Chinese Academy of Sciences, Beijing, China.
His research interests include computer vision, machine learning and remote sensing.
\end{IEEEbiography}
\vspace{-13 mm}
\begin{IEEEbiography}[{\includegraphics[width=1in,height=1.25in,clip,keepaspectratio]{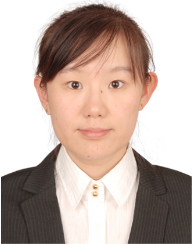}}]{Rui Wang}
Rui Wang received the B.E. degree from Ocean University of China, Shandong, China, in 2009, and the M.E. degree
from Shandong University, Shandong, China, in 2012. She is currently an Senior Engineer with Science \& Technology
on Integrated Information System Laboratory, Institute of Software Chinese Academy of Sciences, Beijing, China.
Her research interests include deep reinforcement learning and multimedia technology and systems.
\end{IEEEbiography}
\vspace{-13 mm}
\begin{IEEEbiography}[{\includegraphics[width=1in,height=1.25in,clip,keepaspectratio]{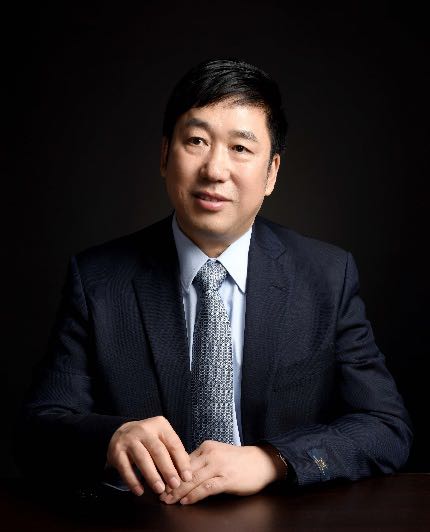}}]{Xiaohui Hu}
Xiaohui Hu received the Ph.D. degree from Beihang University, Beijing, China, in 2003. He is a Professor with
the Science \& Technology on Integrated Information System Laboratory, Institute of Software Chinese Academy
of Sciences, Beijing, China. His research interests include intelligent information processing, big data analytics
and cooperative multi-agent systems.
\end{IEEEbiography}







\end{document}